# A foundation for systematic analysis of transformers and RNNs for tractography

Emmanuelle Renauld[1], Philippe Poulin[1], Hugo Larochelle[2],
Antoine Théberge[1], Maxime Descoteaux[1,3]

[1] Computer Sciences Department, Université de Sherbrooke [2] Mila, Université de Montréal
[3] Imeka Solutions Inc, Sherbrooke, QC, Canada

**Abstract:**

Machine learning (ML) has emerged as a promising approach for improving diffusion MRI (dMRI) tractography, a task that remains limited by the intrinsic tension between local diffusion information and global anatomical plausibility. In this work, we systematically evaluate recurrent neural networks (RNNs) and Transformer models for iterative tractography, with particular attention to training strategies, input representations (including convolutional neural network (CNN)-based embeddings and end-of-sequence (EOS) tokens), and hyperparameter selection. We introduce a generation-validation phase enabling supervision at the streamline level during training, allowing supervision despite the mismatch between local loss functions and global streamline quality. Using the ISMRM2015 tractography challenge dataset, our models achieve the highest reported performance to date. Through controlled experiments, we quantify the impact of missing bundles, noisy or imperfect training streamlines, and invalid fibers in the training set. Finally, we demonstrate the applicability of our best-performing models for *in vivo* data from the Tractoinferno database. Overall, our results highlight both the potential and the limits of sequence-based deep learning models such as Transformers and RNNs for tractography, and emphasize the need for improved phantoms and evaluation methods for in vivo validation. We provide takeaways and recommendations for future researchers training and validating sequence-based supervised methods for tractography.

## 1. Introduction

Machine learning (ML) has become an increasingly valuable tool in diffusion magnetic resonance imaging (dMRI), supporting a wide range of tasks including local reconstruction (Elaldi et al., 2023; S.-M. H. Hosseini et al., 2022; Zeng et al., 2022), segmentation of the white matter (Lu et al., 2022; Peretzke et al., 2023; Xue et al., 2022), segmentation of tractograms into bundles (Labra et al., 2017; Zhang et al., 2022), filtering of tractograms (Legarreta et al., 2020), tractometry (Attyé et al., 2021), connectomics (Bayrak et al., 2019; Sarwar et al., 2020), and more. In this work, we aim to develop an ML-based algorithm specifically for the task of iterative wholebrain tractography from dMRI data.

Tractography is widely considered one of the most challenging steps in dMRI processing. Tractography algorithms require adequately denoised dMRI processed with state-of-the-art techniques to create a local model in each voxel, which is itself a challenge (Jbabdi & Johansen-Berg, 2011; Jones et al., 2013). Then, the tracking process depends on the chosen algorithm. For instance, in iterative processes – arguably the most common – streamlines are generated by advancing step by step from a chosen seed position inside a seeding mask, with a chosen initial direction. The next direction at each step is chosen from the local model and is followed with a chosen step size. Tracking is continued until a stopping criterion is chosen, for instance stepping outside a tracking mask, choosing an implausible next direction (usually one that creates an angle above a chosen threshold relative to the previous direction) or creating a streamline of exaggerated length (e.g., usually above ~200 to 300 mm). Then, the tracking process is repeated from the same seeding point in the reversed direction to track the second half of the streamline. This results in many parameters, all of which have a large influence on the final streamline (Jbabdi & Johansen-Berg, 2011).

But the main reason why tractography is challenging is not only because of these details, but because of inherent limitations (Schilling et al., 2020). There is a duality in the tractography objective: it aims to create streamlines that are both correctly aligned with *local* information (the diffusion signal) and consistent with *global* fiber pathways in the brain, connecting appropriate cortical regions. Due to limitations of medical images such as voxel resolution, noise, or artifacts, it is not possible to fully satisfy both objectives at once. Even in a perfect, noise-free acquisition, the local information in a voxel cannot represent individual fibers (neurons) or even larger-scale fiber pathways (bundles or fascicles of axons), because it reflects the average acquired signal from all underlying structures (Schilling et al., 2020), including axons of varying orientations and some free-water movement (Jelescu & Budde, 2017). When relying solely on local information to produce streamlines, tractography challenges have shown that following the main direction of anisotropy (e.g., deterministic tracking) is often not enough and, while it produces few spurious streamlines, it also results in fewer pathways and poor spatial coverage, as opposed to probabilistic tractography (Maier-Hein et al., 2017; Schilling et al., 2019).

In traditional (non-ML-based) tractography, this global-local duality led to two main types of algorithms: global and iterative. Global tractography encompasses a heterogeneous set of algorithms, but generally the creation of streamlines depends on optimizing a global criterion, which is often difficult both to formulate mathematically and to optimize, limiting the widespread adoption of global tractography. With iterative tractography, algorithms are easier to define: they align each streamline segment on the underlying local diffusion reconstruction model. But the resulting streamlines can easily follow paths that are not supported by any anatomical fiber, as long as the local information supports it (Maier-Hein et al., 2017). Iterative algorithms are particularly prone to generating anatomically implausible streamlines in regions of fiber crossing or bottleneck regions (Schilling et al., 2022). In recent years, the tractography community seems to have been stuck on the problem of streamline accuracy, and only a few improvements have been proposed for iterative algorithms. Adjustments to base algorithms have been suggested to enable better tractography in difficult-to-track regions (e.g., particle filtering tractography (Girard et al., 2014), parallel transport tractography (Aydogan & Shi, 2021), or IFOD2 (Tournier et al., s. d.)), or to include anatomical priors (e.g., through anatomically constrained tractography (Smith et al., 2012)) but it remains unclear how to fully incorporate our knowledge of anatomy into tracking algorithms. This knowledge is intuitive for experts who have developed an “expert eye” through extensive study of specific bundles; for instance, expectations on the path, shape, smoothness, fanning, division into sheets, relationship to surrounding bundles, etc. But this is difficult to inject rigorously into tracking algorithms, except perhaps by performing bundle-specific tracking (BST) for bundles that are already well described in the literature, either with specific rule sets and intensive post-tractography filtering (Chamberland et al., 2017; Rheault et al., 2019; Wakana et al., 2007) or by pre-computing fixel-based maps (Wasserthal et al., 2018). In wholebrain tractography, post-tractography filtering and segmentation are possible (Daducci et al., 2015; Garyfallidis et al., 2018; Pestilli et al., 2014; Smith et al., 2015; St-Onge et al., 2023), but also difficult.

One hope of ML is that it could provide a way to automatically incorporate some knowledge of the general shapes of bundles in the brain, as well as a non-local view of the currently tracked streamline and a general understanding of the organization of the brain, allowing wholebrain tractography with results at least as good as those obtained with BST.

### *RNNs and Transformers: the natural choices for tractography?*

Various models have been proposed for ML-based tractography (see **Figure 1**). In all cases, the output is a direction. In type 1 models, the input consists of the local dMRI data. These models can be viewed as improved local reconstruction algorithms (S. M. H. Hosseini et al., 2022; H. Li et al., 2021; Zeng et al., 2022). By definition, these algorithms cannot outperform classical tractography because they cannot better choose the tracking direction in regions where the local information is either ambiguous (e.g., in bottlenecks (Schilling et al., 2022)) or of poor quality, as they do not have access to any additional global information. These models include random forests (Neher et al., 2017) and neural networks (Wegmayr & Buhmann, 2020).

In type 2 models, algorithms take as input the local information of the N previous points (the local dMRI, possibly with the associated last direction), concatenated. This approach attempts to provide the model with a sense of history, making it more than “just” a local reconstruction model. Although this has proven effective (Neher et al., 2017; Poulin et al., 2017; Théberge et al., 2021; Wegmayr & Buhmann, 2020), it is likely a limited solution: the number of previous directions must be fixed, and the model receives a large input with no explicit representation of the sequential nature of the points.

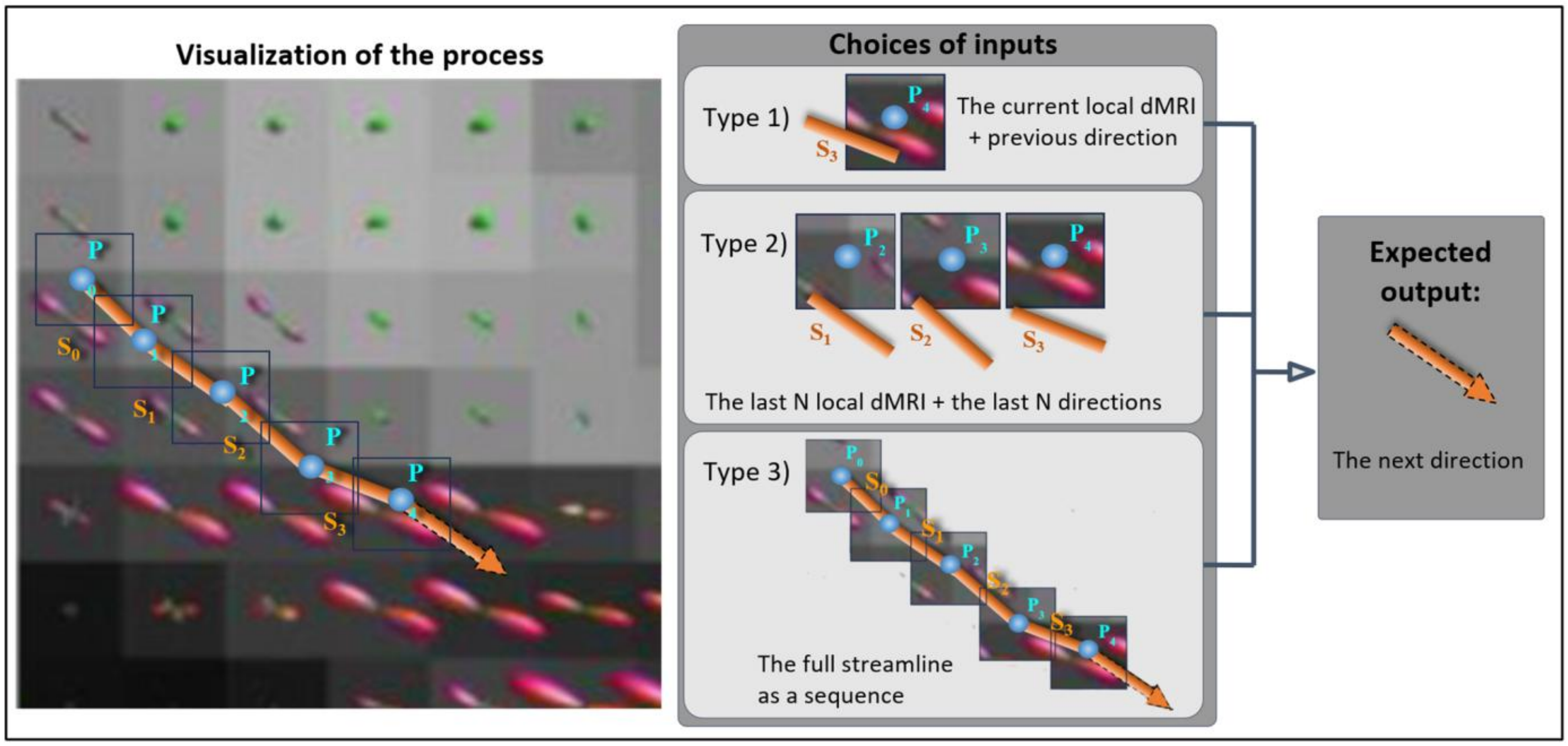

**Figure 1.** Inputs to ML models for tractography. Given points $P_0$ to $P_4$ and segments $S_0$ to $S_3$, the models predict the next direction. Type 1 models use only local features at $P_4$, and optionally the previous direction $S_3$. These models are in fact *not* tractography models but local modeling algorithms. Type 2 models receive the concatenated *N* previous points, providing them with a form of memory. Type 3 models are inherently designed to process sequential data (RNNs, Transformers).

Type 3 models are likely the most suitable for handling streamlines. They are models that are inherently defined to process sequences, such as recurrent neural networks (RNN) or Transformers (Vaswani et al., 2017). In these sequence-based models, some information about the entire streamline is retained (through the hidden state in RNNs; through the attention mechanism in Transformers), making them specifically designed for tractography. RNNs have already been used in Learn2track (Poulin et al., 2017) and in Deeptrack (Benou & Riklin Raviv, 2019), two similar algorithms. Transformers have been used in Tract-RLFormer (Joshi et al., 2024), TractoTransformer (Waizman et al., 2025) and in Yang *et al.*'s work (Yang et al., 2025).

In this work, we further tested both RNNs and Transformers, with careful selection of training sets and hyperparameters. Our code, including both the updated Learn2track (Poulin et al., 2017) and our Transformer model, similar to TractoTransformer (Waizman et al., 2025), is available as the dwi_ml library (https://dwi-ml.readthedocs.io/). Trained models and learned parameters of final experiments are available on Zenodo (10.5281/zenodo.19709562) and can be used with dwi_ml, release 1.1.1.

### *Local loss vs. tractography quality*

In the task of iterative tractography, models are used for a generative task. This makes them inherently difficult to train efficiently, because they are subject to a duality similar to the one discussed above: their loss function is calculated at each point individually, *locally*, but the quality metrics are evaluated on the final generated sequences as a whole, *globally*. Unlike classical tractography, the local-global duality in the ML context does not arise from the fact that voxels represent average signals, but rather comes from the way ML models are trained, namely through the optimization of a local loss function. This difficulty is present in all fields of study that use sequence-based models for generation tasks. During training, the next direction at point N is predicted using inputs from the N-1 previous points of the reference sequence, which comes from the training set. During generation (at test time), however, the model does not have access to a reference sequence, only to its own output (see **Figure 2**). The error accumulates at each new step. In tractography, this means that at each iteration, the generated streamline possibly resembles training streamlines less and less. This issue is known as the exposure bias (Ranzato et al., 2016; Schmidt, 2019).

Two major questions arise from that drawback: 1) Is the local loss a good way to train models, allowing the model to learn the tractography task? 2) Assuming that it is, does it allow validation? And if we cannot rely on the loss itself for a validation criterion, how can we perform an efficient hyperparameter search?

To answer the first question, there is empirical evidence that the local loss does allow models to be trained. Previous studies have achieved good results (Benou & Riklin Raviv, 2019; Dai et al., 2023; Y. Li et al., 2025; Neher et al., 2017; Poulin et al., 2017; Théberge et al., 2021; Waizman et al., 2025; Wasserthal et al., 2018; Wegmayr & Buhmann, 2020; Yang et al., 2025; Zhao et al., 2023), and the local loss has been used effectively in the literature in iterative algorithms, suggesting that it is adequate to help the model learn *something* about tractography. It is difficult to eliminate a loss function entirely as it is the basis of optimization algorithms such as gradient descent. Note that reinforcement learning does remove the need for a local loss function, but this merely shifts the problem to defining a good reward function for tractography (Théberge et al., 2021; Zhao et al., 2023). In the current project, we chose to keep sequence-based models, even though they require a local loss function.

To answer the second question, following the conclusions of recent consensus papers, the local information should not be used for validating tractograms (Schilling et al., 2019). But it is possible to use global scores at validation time. To the best of our knowledge, no paper in the ML-based tractography literature has used a validation set to perform a hyperparameter search. An appropriate validation technique would be to use the model to create a full wholebrain tractogram and to score the tractogram as a whole, at each epoch, provided that a method exists to score whole streamlines. This is however time-consuming and other options can be explored. In all cases, the quality of the generation process must be assessed, not only the quality of the local loss. In this work, we added a generation phase during the validation stage, referred to as the generation-validation (GV) phase in this text, where the generation of complete streamlines allowed better evaluation of the model's quality. We conducted what is, to our knowledge, the most extensive hyperparameter search reported for an ML-based tractography algorithm. Our models outperform state-of-the-art methods on the dataset from the ISMRM 2015 Tractography Challenge (Maier-Hein et al., 2017), referred to as the ISMRM data in this text. We also validated our models *in vivo* on the Tractoinferno database.

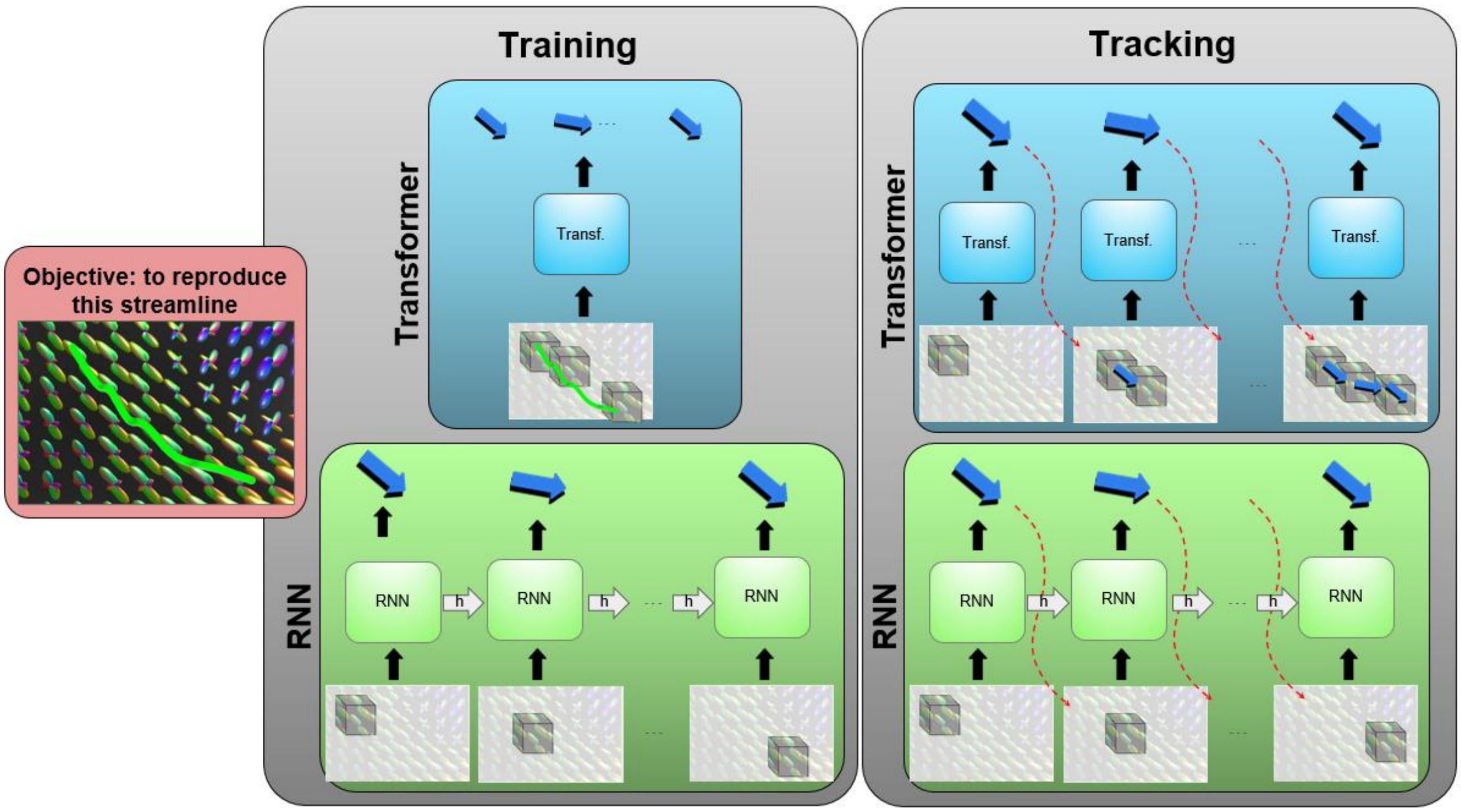

**Figure 2.** Diagram of the training and tracking processes for RNNs and Transformers. At training time, the RNN receives one input at a time in a recurrent fashion, and the Transformer receives the whole streamline at once, making Transformers potentially better at handling long sequences. At tracking time, in Transformers, the entire list of input points is provided to the model at each new time step, making generation with Transformers more computationally expensive than with RNNs.

## 2. Methods

### 2.1 Preparation of the GV phase

Evaluating the generative performance of tractography models would ideally require tracking full tractograms at each epoch, which is computationally expensive and introduces additional design choices (e.g., number of seeds, seeding mask, tracking mask). To address this, we defined a generation-validation (GV) phase. In this phase, for each batch of streamlines in the validation set, we re-tracked streamlines by initializing them from the first $N$ ground-truth coordinates and letting the model generate the remaining segments (in this work, we used $N$ =

1). This procedure allowed us to evaluate generation quality at each epoch while controlling the number of streamlines, making it significantly faster than whole-brain tractography. We then compared the generated streamlines to the reference streamlines.

Rather than relying solely on geometric distances between endpoints, we evaluated whether generated streamlines form valid connections. For the ISMRM dataset, validity could in principle be assessed directly using the available ground truth (Renauld et al., 2023). However, because our experiments on ISMRM data were designed as a preparatory step toward *in vivo* experiments (see experiments description below), where no such ground truth is available, we adopted a unified evaluation strategy across all experiments. Specifically, we considered a streamline valid if it connected regions that are observed to be connected in the subject's validation set, and we compared generated streamlines to the connectivity matrix derived from the validation data.

Our main validation metric is the connectivity false positive (FP) rate, defined as the ratio of false positives to true positives (FP:TP) for endpoint connections. True negative and false negative rates cannot be estimated without generating full tractograms and are therefore not considered here.

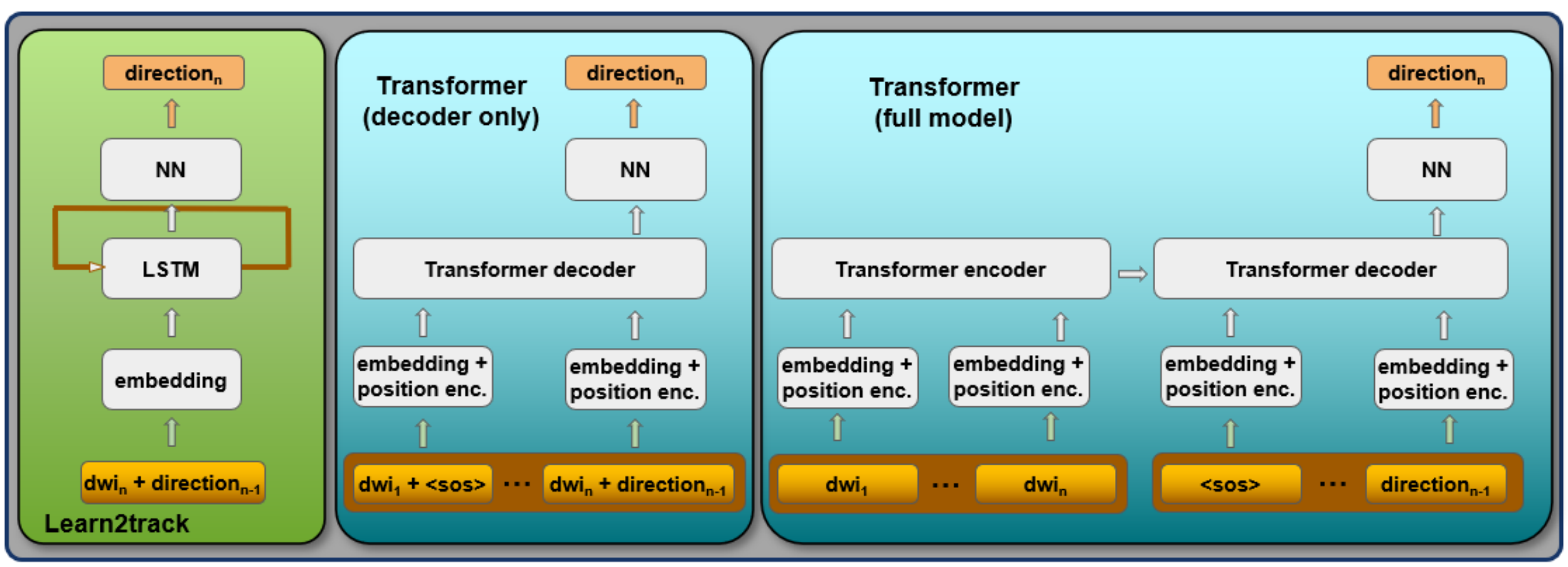

**Figure 3.** Models included in this work. Learn2track (Poulin et al., 2017) uses a RNN. Transformers variants include the full original architecture (Vaswani et al., 2017) or a decoder-only version similar to TractoTransformer (Waizman et al., 2025) or to Yang et al. (Yang et al., 2025). The inputs at each time point include the dMRI neighborhood. In the full-model Transformer, the input to the decoder is the previous direction at each point. Since no previous direction exists at the first coordinate of the streamlines, start-of-sequence (SOS) tokens are added. In the other models, the previous direction can be concatenated to the input. Target direction at the last point does not exist, but can be set as an end-of-sequence (EOS) token. Other details are described in **Table 1.**

## 2.2 Model architecture

RNN models were used with the same architecture as in Learn2track, published in Poulin et al. (2017). For Transformers, two versions of the model were tested (see **Figure 3**). In Transformers models, future timepoints were masked in the attention. The decoder-only version is causal and auto-regressive and a natural fit for tractography, but the full decoder-encoder model can be made causal by masking the attention in the encoder. Using the full model could potentially allow a better decoupling of feature extraction (encoder) and sequence generation (decoder) but is heavier in memory.

**Table 1:** Details of the architecture of the models

| | |
|---|---|
| **Local loss** | Most local losses found in the ML-tractography literature were tested.<br>- Regression: the cosine similarity or the L2-norm between the real and estimated directions.<br>- Classification[a]: the negative log-likelihood.<br>- Smooth classification: the smooth classification suggested by Benou and Riklin Raviv (2019).<br>- Regression of parameters for a Gaussian model: see description in Poulin et al. (2022).<br>- Regression of parameters for a Fisher-von-mises model (FvM): see description in Wegmayr et al. (2020).<br><br>[a] Classes refer to a list of discrete points on the sphere, representing chosen possible directions. |
| **Outputs** | The next direction at each point, formatted based on the choice of local loss function:<br>- Regression of directions: 3D vectors ([x, y, z]).<br>- Regression of Gaussian parameters: 6D vectors ($[\mu_x, \mu_y, \mu_z, \sigma_x, \sigma_y, \sigma_z]$).[b]<br>- Regression of FvM parameters, 4D vectors ($[\mu_x, \mu_y, \mu_z, \kappa]$).[b]<br>- Classification: vectors of length K, where K is the number of points on the sphere.<br><br>[b] Deterministic tracking is performed using the learned mean direction, $[x, y, z] = [\mu_x, \mu_y, \mu_z]$. |
| **Targets** | The targets **Y** are streamlines formatted as sequences $[\mathbf{y}_1, \mathbf{y}_2, \ldots, \mathbf{y}_{n-1}]$ of directions: the difference between consecutive coordinate points. In the case of classification, outputs were formatted as one-hot vectors representing the discrete directions on the sphere. For smooth classification as described in Benou and Riklin Raviv (2019), the one-hot vectors were multiplied by a smoothing Gaussian. Target at the last point of each streamline was either ignored as no next direction exists there, or added as an end-of-sequence (EOS) token (see below). |
| **Inputs** | The inputs **X** were sequences $[\mathbf{x}_1, \mathbf{x}_2, \ldots, \mathbf{x}_n]$ of MRI data at each coordinate: either the raw data or the spherical harmonics (SH) parameters for the fODFs, possibly concatenated with the T1 or the fractional anisotropy (FA) value at that point. fODFs were saved as spherical harmonics using the Descoteaux07 basis (45 features). Data was normalized, using data = (data – mean)/std. Mean and standard deviation (std) were computed in the brain mask, to avoid including the background, which would bring the mean and std very close to 0. If many MRI modalities were used (ex, fODF, T1, FA), normalization was performed separately on each volume before the concatenation of features.<br>Optionally, the preceding direction could be added as another input feature (see **Figure 3)** at each point, either as [x, y, z] directions or as classes. Then, the preceding direction at the first |
| | coordinate was added as a SOS token (see below). If no EOS token was used (see below), the last point $\mathbf{x}_n$ was rejected from data.<br>In Transformers, positional encoding on the positions 1 to *n* used the sinusoidal encoding like in the original paper (Vaswani et al., 2017), with a default maximum length of 1000, which was enough to cover all streamlines. |
| **EOS / SOS** | Optionally, Start of Streamline (SOS) and / or End Of Streamline (EOS) tokens could be added. The SOS / EOS format depended on the data format of the input / target directions, respectively.<br>- Directions as [x, y, z] coordinates: we added an additional value f to the vectors ([x, y, z, f]), where f was a flag with value 1 at the endpoint, and 0 everywhere else.<br>- Directions as classes: the SOS/EOS represented an additional class. |

Also, contrary to what is typically seen with Transformers, future inputs were also masked: the dMRI information at future points depend on the newly generated coordinates of the streamline. We evaluated both the original encoder–decoder Transformer architecture (Vaswani et al., 2017) and a simplified decoder-only variant, where the dMRI information and the target at each past point were concatenated together as input.

The common architecture for all our models is detailed in **Table 1**.

## 2.3 Experiments

Experiment 1: Hyperparameter search. We used the ISMRM data to perform an analysis of the hyperparameters (2mm isotropic diffusion acquisition, with 32 gradient directions, b-value=1000 s/mm2, and two b0 volumes, one with reversed phase-encoding direction). Due to the small size of the dataset, training on the ISMRM data was fast, and this allowed an extensive analysis of parameters before testing our models on a bigger database. For resemblance with real data, the dMRI volume was taken from the noisy dataset rather than from the ground-truth (GT) unnoisy dMRI. We used the preprocessed data published in Renauld et al. (2023) and available at www.tractometer.org. This data was preprocessed with Tractoflow (Theaud et al., 2020) version 2.3.0. Inputs included the second b0 as a reversed b0 to allow the topup step. The processed data include the denoised and registered T1, the FA map, the local fiber orientation diffusion functions (fODF), and a denoised version of the dwi (from the Resample_DWI folder). As targets, we used the GT streamlines for this experiment. Available streamlines were randomly split into two subsets for training and validation, and each subset was downsampled to 70 000 streamlines. For the connectivity matrix, high-quality parcellation of the gray matter proved difficult to obtain. Instead, we separated the full space into 6 x 6 x 6 blocs and used each one as a separate region of interest (ROI) (see **Supplemental Figure S1**).

We tested an extensive range of parameters, including the formulation of the direction and associated loss (see **Table 1**), the segments length in streamlines (step size during tractography is then the same), the use of an EOS token, the type of data embedding (neural networks [NN] or convolutional neural networks [CNN]), PyTorch optimizer, the number of layers in models, the dropout rate, the batch size, the use of weight decay, the number of neighbours interpolated in the dMRI volume. We tested the addition of noise to the streamlines as a means of data augmentation. We also investigated the effect of excluding the FA and T1 from the input, or of replacing the fODF with denoised dMRI data. For Learn2track, additional parameters included the use of layer normalization and skip connections. For TractoTransformer, additional parameters included the choice of model (see **Figure 3**), the option of moving the normalization layer to before the attention sublayers such as suggested elsewhere (Xiong et al., 2020), and the choice of formatting for the SOS (see **Table 1**). During training and validation, half of the streamlines were reversed on-the-fly, at each epoch, for data augmentation. Training was performed using PyTorch. For each hyperparameter configuration, training was monitored using the FP:TP rate of the GV phrase. Configurations were retained when they resulted in consistent improvements in validation scores over the full training process, rather than improvements attributable to isolated fluctuations in the learning curve. Models were trained with a patience of 50 epochs without improvement.

For the GV phase, the tracking mask was the GT white matter (WM) mask: the mask of all voxels touched by a GT streamline. The maximal length was 200mm.

After models were trained, they were used for whole brain tractography as a test phase. The tracking mask was again the WM mask, length range 10 to 200mm. The maximal angle between two consecutive steps was relaxed compared to typical values (ex, 45°) and set to 90° to allow better perceiving the models' tracking ability. Other tracking choices were set to reflect the model's architecture: A) step size was the same as segments length in input sequences during training. B) The seeding mask was set as the interface mask: the mask of voxels including an endpoint of a GT streamline. This reflects the learning process; our models always read streamlines from the beginning. In Learn2track's initial form, published in Poulin et al. (Poulin et al., 2017), the streamlines were split into two at a random point during training, to ensure that the model could start from anywhere in the white matter. We argue this complicates the discovery of anatomical priors and we chose to use complete sequences only. Importantly, a full interface mask could have been better to analyse the model's quality, but was difficult to produce robustly in the ISMRM data. We used 1 seed per voxel. We used deterministic tracking for all models, even for models that would allow probabilistic tracking (Smooth classification, Gaussian, Fisher-von-Mises). The final metrics for the testing phase were the ratio of invalid streamlines (IS) and the F1 score, using the reappraised scoring system suggested in Renauld et al. (2023). The IS score is sometimes preferably discussed as the ratio of valid streamlines (VS), and IS + VS = 1. The F1 score is also known as the Dice score, and can be seen as a measure of the compromise between the overlap (OL) and the overreach (OR). The overreach is generally presented as a ratio of the GT bundle's volume, and then called $OR_{gt}$.

The main metric of the GV phase (the FP:TP rate) was chosen as a means of improving the speed of the training / validation phase, but is arguably flawed as it could incorrectly score streamlines ending in an appropriate region but with an invalid overall pathway. To evaluate the pertinence of the FP:TP score, as supplementary analysis, we investigated the correlation between the GV phase's score and testing phase's F1 score throughout epochs. To investigate the ability of the model to correctly produce EOS probabilities, tractography was re-computed without a tracking mask (a simple brain mask, i.e. a skull-stripped mask, was used to ensure streamlines were stopped if they left the brain, reducing tractography computation time).

Experiment 2: Towards real data part 1 – The ability to recover missing data. With *in vivo* data, there is no guarantee that all existing bundles are present in the training set. In preparation for our work on a bigger database, we analyzed our models' ability to recover bundles that were not seen during training. We modified the training set, discarding one bundle from the ISMRM data. We verified if, at test time, that bundle was recovered with tractography. We tested the effect of discarding a) the biggest bundle, the corpus callosum (CC), b) an intermediate bundle, the superior cerebellar peduncle (SCP_left), and c) a small bundle: the anterior commissure (CA). These three bundles also each belong to a different category of bundles based on Maier-Hein et al. (2017): medium difficulty for the CC, hard-to-track for the SCP, and very hard-to-track for the CA. To investigate the result, we performed whole-brain tracking as in Experiment 1 (number of seeds per voxel=1), and also performed a more extensive tracking starting only from the endpoints of the GT missing bundle (number of seeds per voxel = 10).

Experiment 3: Towards real data part 2 – The ability to train on sub-par streamlines. In preparation for our work on a bigger database, we investigated the effect of using "gold standard" streamlines as targets instead of ground truth streamlines, which are not available *in vivo.* We used an in-house tractogram obtained from classical tractography on the ISMRM data, rather than the GT streamlines. We used scil_tracking_local from the scilpy library (Renauld et al., 2026). We tracked inside the GT WM mask, seeding again from the GT interface mask, with 10 seeds per voxel, maximal angle theta 45°, step size 0.5mm, minimum length 10mm, maximum length 200mm. We tested three datasets: deterministic tracking, probabilistic tracking and a combination of both. When using both, during training, the number of streamlines per epoch was limited in the same range as the total number of streamlines in the deterministic and probabilistic tracking. With these parameters, we noted that the CA and CP, the two most difficult bundles, were not recovered with classical tractography. To make sure that they would be included in the training data, we ran bundle-specific tracking (BST), using the CA and CP's GT pathway as seeding mask and a twice-dilated tracking mask, with 30 seeds per voxel, and we succeeded to obtain streamlines for these bundles. We combined the results to the first tractograms. Finally, we scored the results (Renauld et al., 2023) and used the recovered valid bundles as the training set. Some bundles were resampled to ensure that they were not over-represented in the training set. Similar to experiment 1, we split the tractogram into training and validation sets. To allow comparison with experiment 1, the GT connectivity matrix was retained for the GV phase rather than using the connectivity matrix from this new data.

Additionally, in preparation for *in vivo* data, for which no ground-truth scoring system exists,

we investigated the relevance of the angular error (AE) as an additional score. The angular error was computed as the mean angle between each local direction of the segments of the streamlines and the direction of the most closely aligned fODF peak[1] in the underlying voxel (with nearest-neighbor interpolation). To investigate the relevance of the AE score for models trained with the ISMRM data, we compared AE scores with F1 scores throughout all models.

Experiment 4: Towards real data part 3 – The ability to ignore false positives. In preparation for our work on a bigger database, we investigated the effect of using invalid streamlines in the training data, mimicking *in vivo* bundles, where there is no guarantee that each streamline corresponds to the underlying anatomy. On the ISMRM data still, using the probabilistic tractogram as targets, we added back 5%, 10%, 20% or 50% of the streamlines that had originally been scored as invalid streamlines (IS) and discarded from the in-house probabilistic tractogram. Training sets were named IS1, IS2, IS3, IS4, respectively. We verified how the final scores were impacted.

Experiment 5: Results *in vivo*. We trained models using the Tractoinferno database (Poulin et al., 2022). We used a subset of the 284 subjects to allow faster processing. We chose 18 training subjects and 6 validation subjects, sampled to maximize the quality of the training data (average number of bundles kept after QC: 30 for the training set, 28 for the validation set), while also keeping a good representation of the acquisition sites. One site was rejected on purpose from the training set and validation set, but added to the testing set, to allow evaluating the generalization capabilities of models, similar to what was done in Poulin et al. (2022). We chose 3 testing subjects: one subject with a high success rate with classical tractography (sub-1178, 30 VB out of 39 kept after QC), one subject with low success rate with classical tractography (sub-1046, 19 VB after QC) and one subject from the rejected acquisition site (sub-1222, 27 VB after QC).

As input, we used the T1, FA and fODF, all openly available. As target streamlines, we used the subjects' valid bundles after segmentation with RecobundlesX (RBX), such as preprocessed by the Tractoinferno team. However, in the official Tractoinferno database, only parts of the recovered bundles are available; some bundles were rejected as part of the Quality Control (QC). Considering that experiment 2 revealed that missing bundles have a high impact on the results, we preferred to use as many bundles as possible and we used all bundles, even if they did not pass the QC by the Tractoinferno authors[2]. We resampled the bundles to 3000 streamlines per bundle to avoid over-representation of easy-to-track bundles. We prepared the connectivity matrices as follows: for each subject, we parcellated the T1 with Synthseg (Billot et al., 2023) and dilated the labels to cover the gray matter (GM). We concatenated the target streamlines together, but rejected streamlines that did not connect two GM regions (1% of streamline on average). We computed the connectivity matrix from the position of endpoints of streamlines.

[1] These fODF peaks are available in the Tractoinferno database and were created with Tractoflow (Theaud et al., 2020), version 2.1.1.

[2] Bundles rejected by QC are not available openly but can be provided upon request.

During training, we used hyperparameters chosen with experiment 1 and trained models with a patience of 15 epochs without improvement.

As a test phase, we analysed the quality of reconstructed tractograms using visual inspection. Current technologies are not precise enough to allow quantifying the results (see more about this in the discussion section). *In vivo* validation remains an open problem. To allow presenting our results, we still segmented our tractograms into bundles to allow visual analysis, as is often done elsewhere in our field. However, we emphasize here that this is not ideal and should be perceived as exploratory. To visualize results using segmented bundles, large segmentations including as many streamlines as possible offer the best perception of results. We used bundles segmented with BundleSeg (St-Onge et al., 2023) , using the Population Average Atlas version 3.1 (https://zenodo.org/records/10103446), with thresholds set manually, as permissive as possible to allow including as many streamlines as possible while still limiting the number of spurious streamlines included.

## 2.4 Analysis of the Transformer's attention

We analyzed the Transformers attention weight in *in vivo* data. We used Tractograms generated by TractoTransformer in experiment 5 for subject sub-1178. Multiple visualization choices were tested: 1) we observed the matrices representing the weights for one streamline at the time, similar as what is done in (Yang et al., 2025); 2) we used the Bertviz toolbox (Vig, 2019), one streamline at the time, 3) we duplicated streamlines at each length and colored them based on the attention weight: for instance, for a streamline $s$ of length $n$, the duplicate $s_i$ with $i \leq n$ is a streamline containing the $i$ first data points of $s$, and the color at each point reflects the weight of the model at each of these points when predicting point $i+1$. Then, it is possible to view the weights as a color. We used MI-Brain (https://github.com/imeka/mi-brain) to view the colored streamlines. We also computed a map of the average weight in each voxel, i.e. the average of the weights at that coordinate for every streamline crossing that voxel. In all cases, attention weights were rescaled to allow easier interpretation of values (see details in **Supplemental Figure S2**).

# 3. Results

Experiment 1: Hyperparameter search. Overall, 177 models were trained. Typically, between 50 and 400 epochs were required, with total training time ranging from 1 to 3 days. Experiments were run on Alliance Canada clusters using one GPU node (either NVidia V100SXM2 nodes with 16G memory or H100 SXM5 nodes with 20G memory). At inference time, tractography was computed on a local workstation with 11 GB GPU (NVIDIA GeForce RTX 2080 Ti). With the final hyperparameter sets, tracking for 60144 seeds took 3:45 minutes with Learn2track (750 streamlines per batch) and 11:00 minutes with TractoTransformer (1500 streamlines per batch).

Examples of GV phase's FP rate learning curves are shown in **Supplemental Figure S3,** and a description of the hyperparameter selection process is detailed in **Supplemental Table S1**. Best scores were achieved with regression for both models, with scores for Learn2track slightly stronger. We detail the final chosen parameters in **Table 2**. For TractoTransformers, final hyperparameters were similar, and the chosen model was the decoder-only version.

**Table 2.** Hyperparameters for the best Learn2track model

| **Parameter** | **Value** |
|---|---|
| Output type | Regression |
| Input | fODF + T1 + FA<br>Previous directions were not explicitly given as inputs |
| Input embedding type (NN or CNN) | NN; output size = 500 |
| Streamlines resampling | Segment length = 2mm |
| EOS token usage | ON |
| Dropout rate | 0.1 in RNN layers<br>0 in last NN layers |
| Number of neighbors | 124<br>(26 voxels at 1mm,<br>98 voxels at 2mm) |
| Noise on streamlines | ON<br>(random Gaussian noise of standard deviation 0.01) |
| RNN unit | LSTM |
| RNN layers and sizes | 3 layers of size 1000 |
| Layer normalization | ON |
| Skip connections | ON |
| Learning rate | 0.001 for 30 epochs, then 0.0005 |
| Weight decay | 0 |
| Optimizer | Adam |
| Batch size | 500 streamlines |

Final epochs' connectivity FP rates from the GV phase are shown in **Figure 4.** For each category of model (Learn2track, TractoTransformer) and output type (classification, regression, etc), the best model's tractography scores are shown in **Table 3**. The best results were obtained with Learn2track, using a regression loss, but all models outperformed previous work. Comparison with previous work is shown in **Figure 5**. Results with the best model were further investigated visually, **see Figure 6**.

As supplementary analysis, the relationship between various metrics are shown in **Figure 7**. We observed that the local loss does not explain well the generative ability as measured by the GV phase (the FP rate), particularly during initial epochs when the local loss is still high. We also observed that the FP rate is well aligned with the IS score.

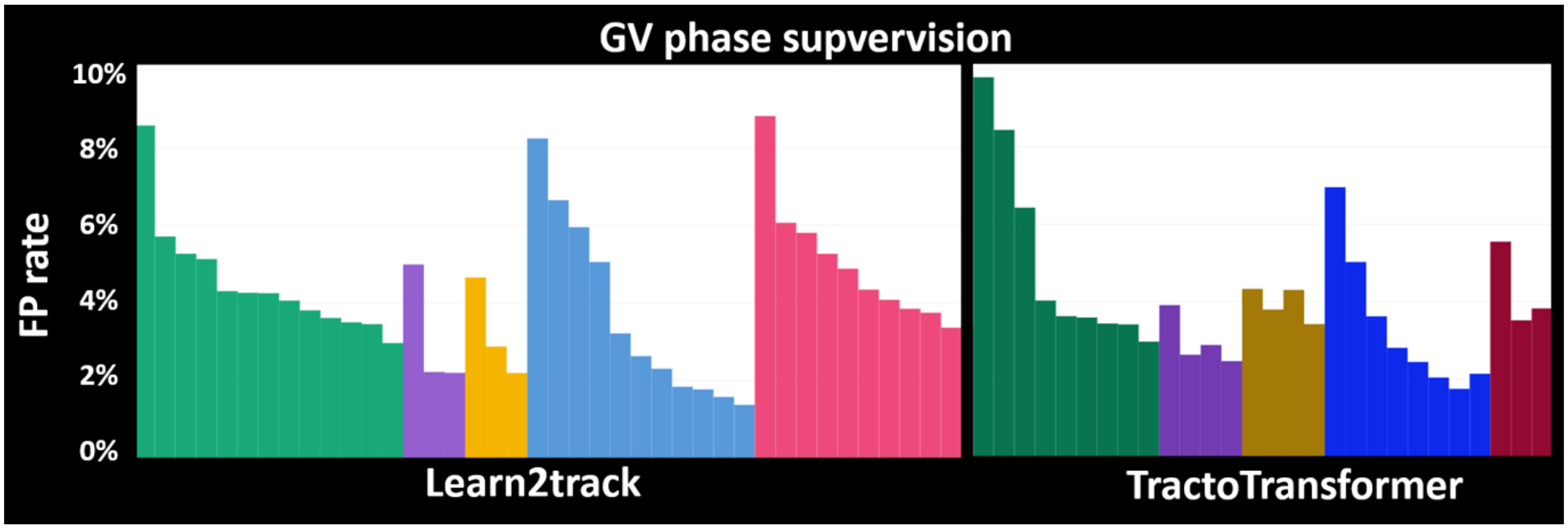

**Figure 4.** Experiment 1: GV phase's connectivity FP rate at the best epoch, for 100 chosen models out of all 177 hyperparameter sets tested. Colors represent the choice of output type. Green: classification; Purple: FvM; Orange: Gaussian; Blue: regression; Red: smooth classification. In all cases, the first modified parameter was the addition of EOS and led to a major improvement.

**Table 3.** Results of experiment 1. Results are presented for generation on both the ground truth white matter mask, or on a full-brain mask using the EOS with a probability of 0.5 as additional stopping criterion. Results with the original scoring system are available in **Supplemental table S2**.

| | | | **Results when tracking in the WM mask** | | | | | **Result when tracking in the brain mask, using the EOS** | | | | |
|---|---|---|---|---|---|---|---|---|---|---|---|---|
| **Training set** | **Model** | **Loss** | **VB** | **VS** | **F1** | **OL** | **$OR_{gt}$** | **VB** | **VS** | **F1** | **OL** | **$OR_{gt}$** |
| ISMRM (GT) | Learn2track (ours) | Classification | 21 | 95.2 | 90.2 | 90.5 | 10.1 | 21 | 98.0 | 87.9 | 93.4 | 18.6 |
| | | **Regression** | 21 | **96.5*** | **91.0*** | **91.3*** | **9.3** | 21 | **98.9*** | **88.4*** | **93.8*** | **18.6** |
| | | Smooth class. | 21 | 94.4 | 90.4 | 90.8 | 10.1 | 21 | 98.6 | 87.9 | 93.6 | 19.9 |
| | | Gaussian | 21 | 95.4 | 89.6 | 89.4 | 10.0 | 21 | 98.2 | 86.7 | 92.5 | 20.6 |
| | | FvM | 21 | 95.0 | 90.3 | 90.9 | 10.4 | 21 | 98.3 | 87.4 | 93.5 | 20.6 |
| | Transformer (ours) | Classification | 21 | 95.1 | 89.8 | 90.3 | 10.7 | 21 | 98.0 | 87.3 | 93.0 | 21.1 |
| | | **Regression** | **21** | **96.1** | **90.4** | **89.9** | **8.9*** | **21** | **98.8** | **88.0** | **92.7** | **18.3*** |
| | | Smooth class. | 21 | 94.4 | 89.4 | 89.7 | 10.8 | 21 | 98.4 | 87.2 | 92.7 | 20.0 |
| | | Gaussian | 21 | 94.1 | 87.3 | 85.6 | 10.0 | 21 | 97.7 | 84.1 | 89.1 | 22.6 |
| | | FvM | 21 | 94.9 | 89.4 | 88.8 | 9.6 | 21 | 98.2 | 86.9 | 91.7 | 19.5 |
| HCP | Transformer (Yang et al., 2025) | Weighted mean squared error | 21 | 64.4 | 59.2 | 65.2 | 48.3 | - | - | - | - | - |

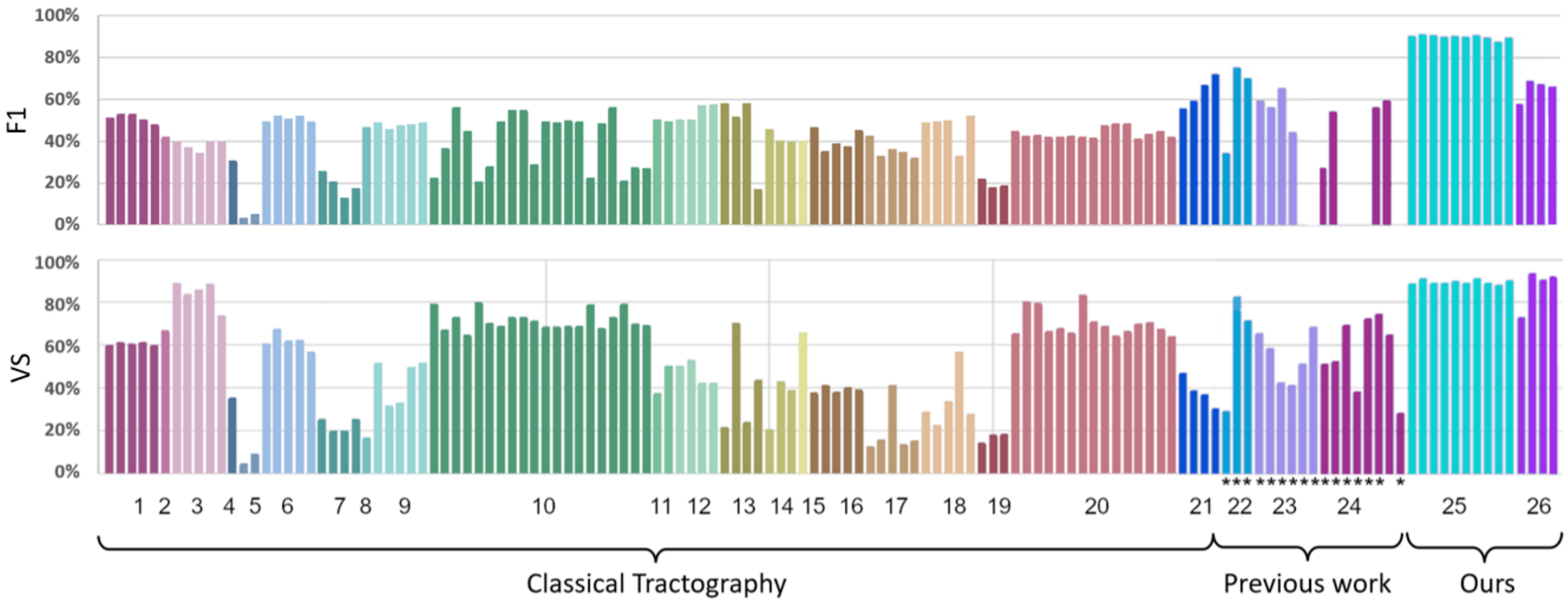

**Figure 5**. Experiments 1 and 3: Results on the ISMRM data, comparison of our results with previous work. Entries with stars below were scored with the original scoring system (Maier-Hein et al., 2017), others with reappraised version (Renauld et al., 2023). For detailed scores, see **Supplemental Table S2.** Some F1 scores were not detailed in previous work.

- Teams 1 to 20: the original participating teams to the ISMRM Tractography Challenge 2015 (Maier-Hein et al., 2017).
- Team 21: Our training sets for experiment 3 (Det or Prob, with or without additional BST for the CA and CP).
- Team 22: Previous work trained on the ground truth (GT) streamlines (Benou & Riklin Raviv, 2019; Waizman et al., 2025; Wegmayr et al., 2019).
- Team 23: Previous work trained on the ISMRM data, but not on the GT streamlines (ex, in-house tractograms or non-supervised work) (Benou & Riklin Raviv, 2019; Poulin et al., 2017; Théberge et al., 2021; Wegmayr & Buhmann, 2020).
- Team 24: Previous work trained on HCP data (Y. Li et al., 2025; Neher et al., 2017; Théberge et al., 2021; Wegmayr et al., 2018, 2019; Wegmayr & Buhmann, 2020; Yang et al., 2025).
- Team 25: Our results from Experiment 1 (trained on GT), also presented in **Table 3**. These models largely outperform previous work from all categories, particularly regarding the F1 score.
- Team 26: Our results from Experiment 3 (trained on in-house tractograms). These models achieve F1 scores similar to their training set (Team 21), but with higher VS ratios. They achieve slightly better F1 scores than previous work also not trained on GT streamlines (Teams 23 and 24).

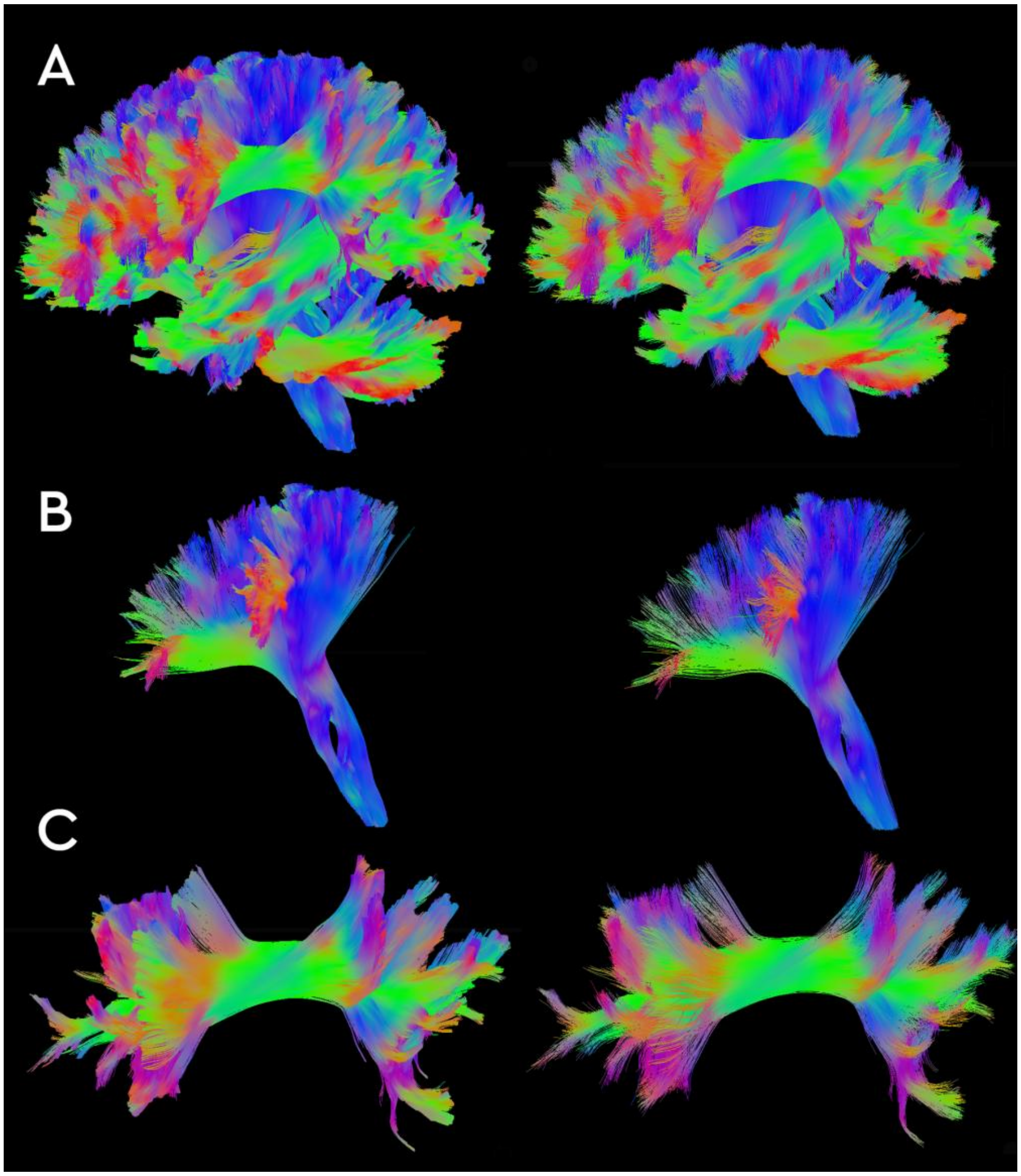

**Figure 6.** Visual inspection of the tractogram generated by the best model in experiment 1: Learn2track + regression. On the left: the GT streamlines (the training set). On the right: The tractogram generated by Learn2track. A. Full brain. B. The left brainstem projection system (BPS bundle). C. The left superior longitudinal fasciculus (SLF bundle). Results are very similar in both cases, except less dense in the tractograms generated by Learn2track. This could be resolved by increasing the number of tracked streamlines.

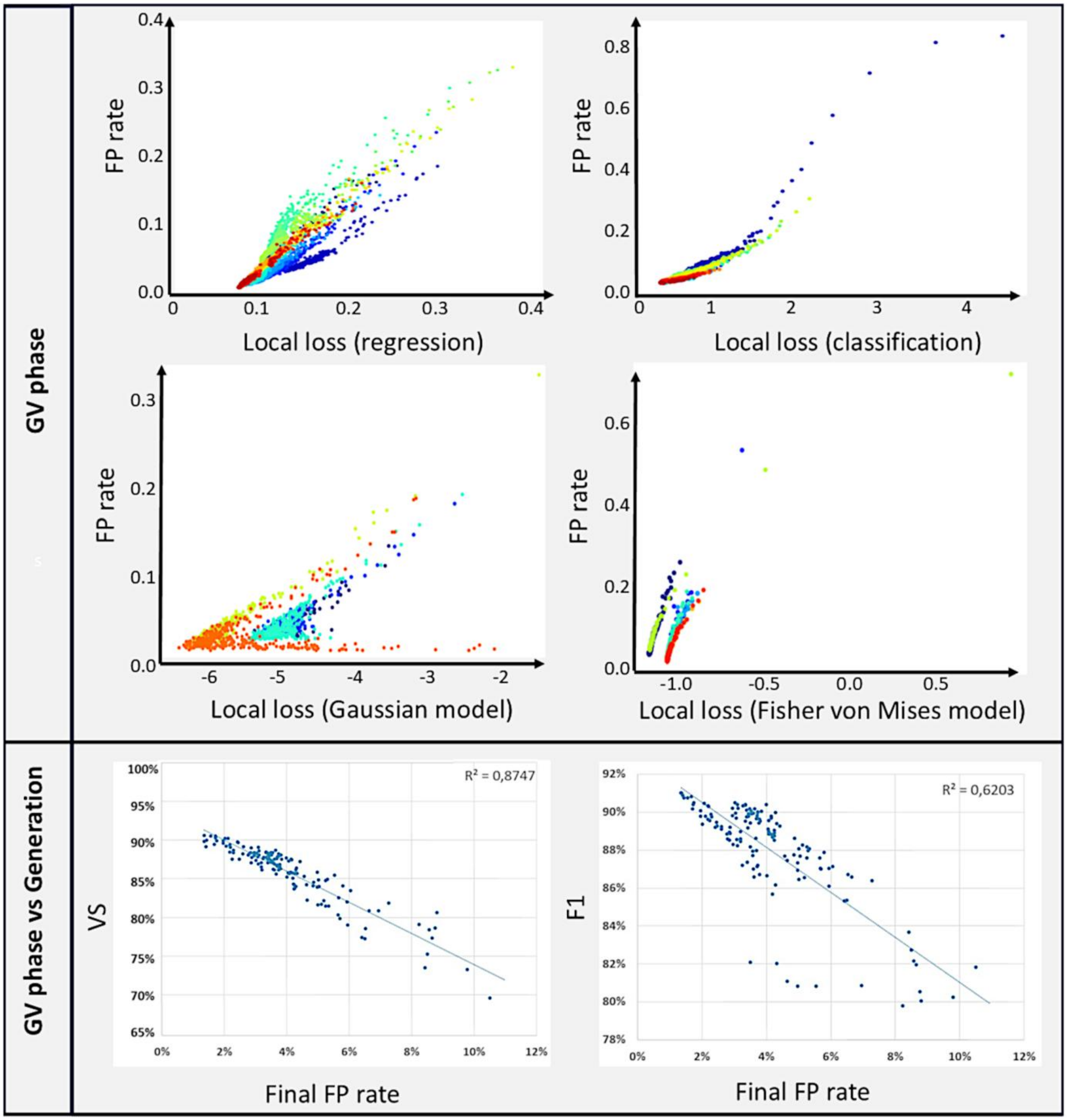

**Figure 7.** Relationship between various metrics. Top row: the GV phase's connectivity FP rate as a function of the local loss. Each point represents an epoch, each color represents a chosen hyperparameter set. Best models have low local loss and low FP rate. Bottom row: the FP rate at best epoch compared with tractometry scores on the tractogram generated with the 177 fully trained models. Best models have low FP rate and high VS, high F1 (top left corner).

Experiment 2: recovering missing bundles. Whole-brain results for Experiment 2 are shown in **Table 4**. When training with a missing bundle from the training test, at generation time the bundle of interest was either not reconstructed or poorly reconstructed. Extensive tracking from the missing bundle's endpoints is shown in **Figure 8.**

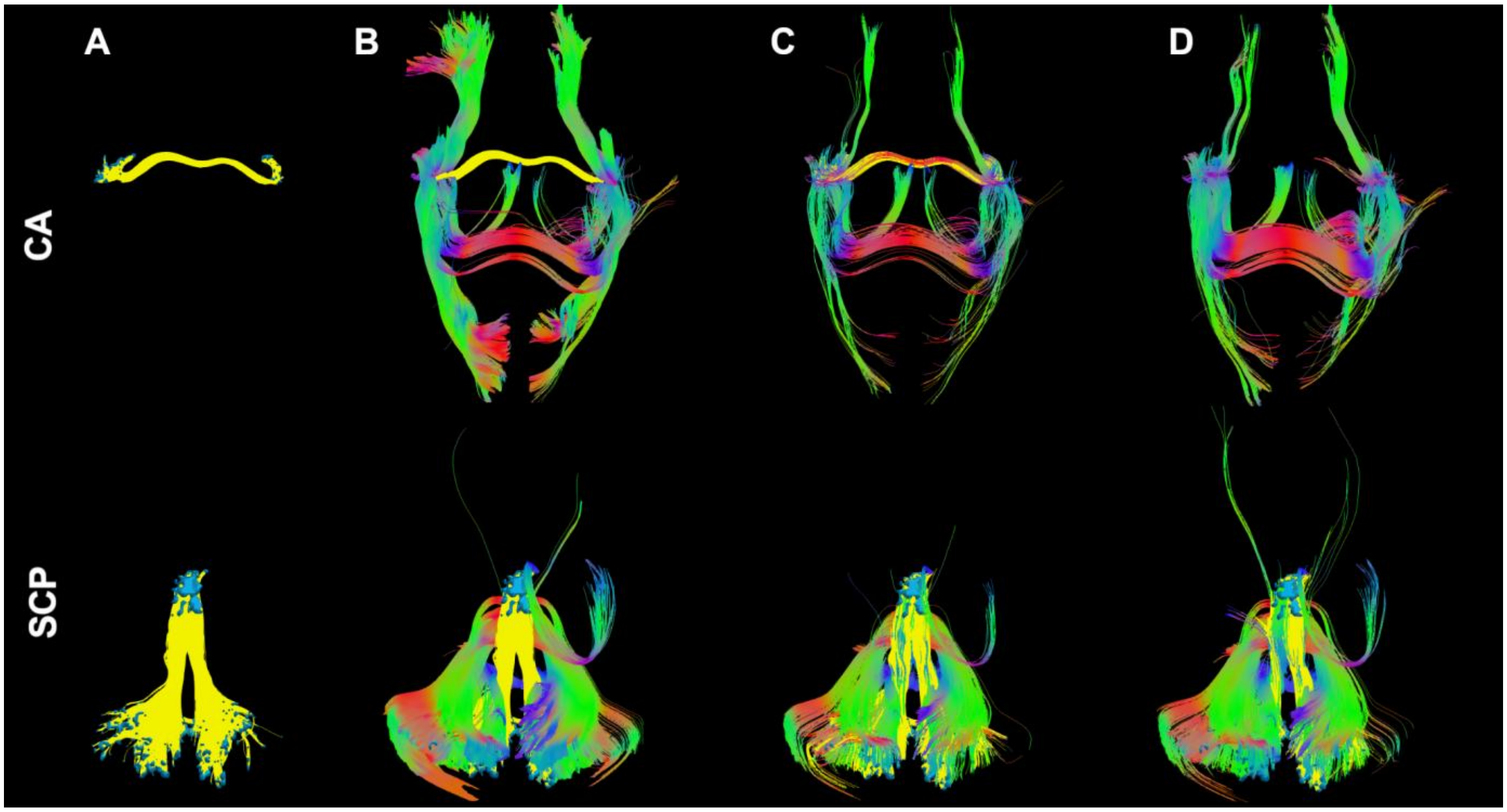

**Figure 8**. Investigation of Experiment 2 results. A) The GT bundles (yellow; the CA in the top row and the SCP on the bottom row) and their endpoints (blue), which become the new seed points. B) All GT streamlines touching the seed points at least once. C and D present tractograms when seeding extensively from these seed points (10 seeds per voxel). C) Using the reference Learn2track model, which was trained on all bundles. D) Using the model that was trained with a missing bundle (top: missing CA, bottom: missing SCP). The CA is not recovered anymore, and the SCP is recovered more partially than in C.

Experiment 3: Training on deterministic and probabilistic streamlines. Results for Experiment 3 are detailed in **Table 5** and are included in **Figure 5** for comparison with previous work. Best results were with a training set generated with probabilistic tracking. Although the VS ratio remained high as compared to training on GT streamlines, the F1 score decreased, driven both by a decrease in OL and an increase in OR. The quality of reconstruction mirrored training data: determinist tracking results in less overreach but also less overlap than probabilist tracking, with streamlines better aligned on the local information (better AE).

When training on probabilistic tractography, the learning curve revealed stronger overfitting of the local loss. To investigate where the validation error was high, we colored the training streamlines and validation streamlines with the local loss (see **Figure 9**). It revealed that high error points were mainly situated on validation streamlines presenting variations compared to training streamlines.

Experiment 4: Adding invalid streamlines to the training set. Results for Experiment 4 are shown in **Table 6.** While the ratio of valid streamlines decreased when invalid streamlines were added in the training set, the F1 score was less impacted, partly because the OR improved.

**Table 4.** Experiment 2: Scores when removing bundles from the training set. Missing bundles were either not recovered or poorly recovered.

| Training set | VB | F1 | | | |
|---|---|---|---|---|---|
| | | CA | SCP_left | CC | Average of the other bundles |
| All streamlines | 21 | 85.6 | 87.4 | 86.3 | 90.6 |
| No CA | 20 | 0.0 | 86.4 | 90.1 | 91.0 |
| No SCP | 21 | 87.2 | 24.9 | 89.8 | 90.7 |
| No CC | 21 | 87.7 | 88.3 | 4.1 | 89.6 |

**Table 5.** Experiment 3: Test-time results by training set, compared with training-set scores. GT denotes ground-truth streamlines (as in Experiment 1). Trends mirror the training data: deterministic tractography generates less overreach (OR) and smoother streamlines better aligned with the local information (better AE), but probabilistic tractography offers better coverage (OL). All training sets included only the valid streamlines (%VS = 100 for all training sets).

| | Training set | | | | | | Results when tracking in the WM mask | | | | | |
|---|---|---|---|---|---|---|---|---|---|---|---|---|
| Name | VB | VS | F1 | OL | $OR_{gt}$ | AE | VB | VS | F1 | OL | $OR_{gt}$ | AE |
| GT | 21 | 100 | 95.3 | 91.1 | 0 | 11.4 | 21 | 95.8 | 90.3 | 90.1 | 9.2 | 11.2 |
| VB-Det | 21 | 100 | 57.3 | 45.4 | 10.8 | 4.7 | 20 | 72.6 | 57.5 | 47.4 | 10.4 | 5.7 |
| VB-Prob | 21 | 100 | 70.4 | 69.0 | 27.7 | 17.2 | 21 | 93.1 | 68.7 | 62.4 | 17.8 | 6.4 |
| VB-Both | 21 | 100 | 70.7 | 70.8 | 30.3 | 10.2 | 21 | 90.0 | 67.0 | 59.7 | 15.9 | 5.9 |

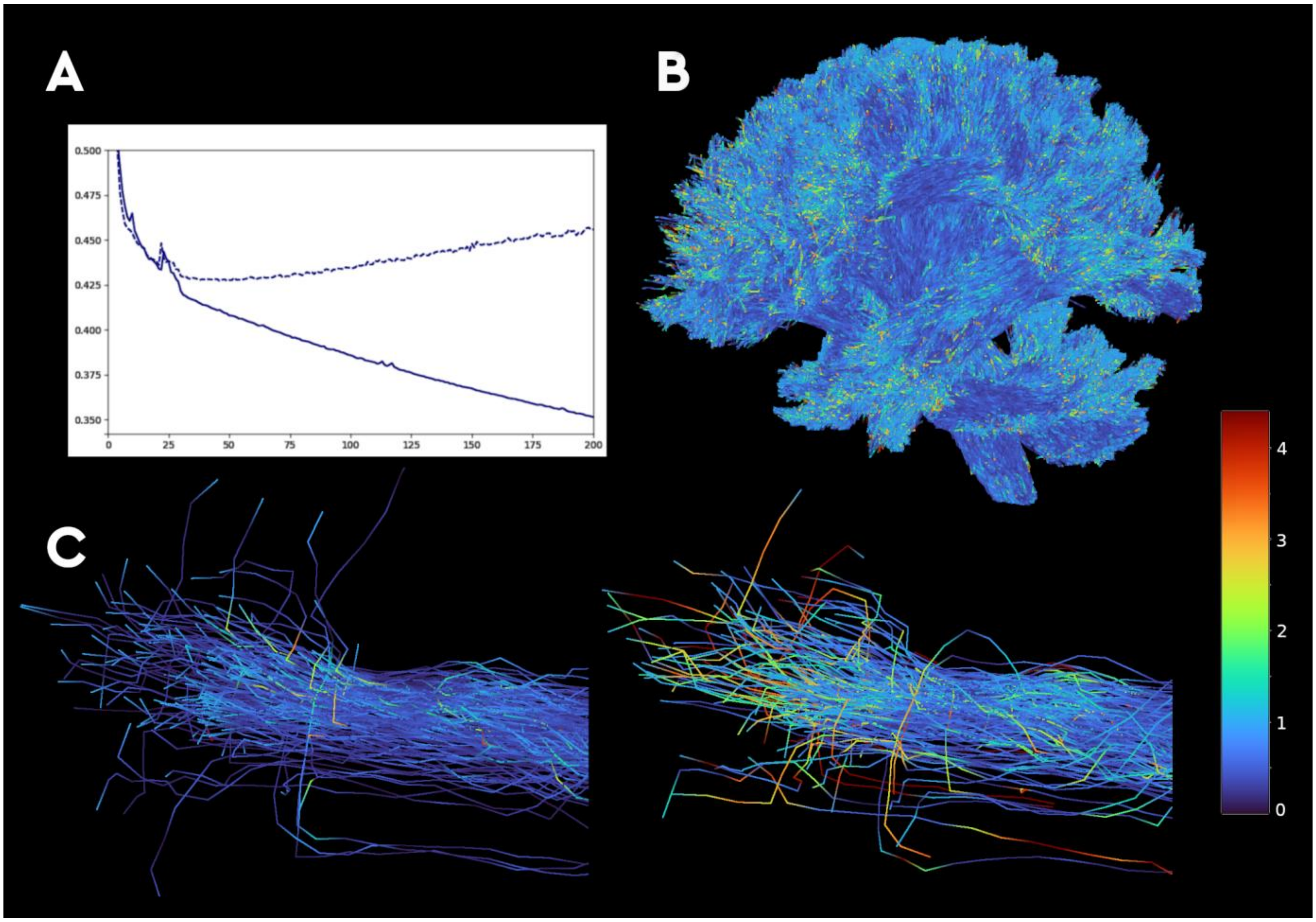

**Figure 9.** Example of overfitting of the local loss. **A.** Learning curve during training on the Prob dataset of the Learn2track + regression model. Solid line: training loss. Dashed line: validation loss. **B** and **C.** Visualisation of the local loss (regression on the direction summed to the binary cross entropy of the EOS) as a color. In B: the full tractogram in the validation set. In C: a zoomed section, comparing the local loss in the training set (on the left) and in the validation set (on the right).

**Table 6.** Result for Experiment 4; adding invalid streamlines to the training set. The number of valid streamlines (VS) and general quality of the spatial coverage (F1) both decreased when invalid streamlines were added, but scores stayed similar to results generally obtained with classical tractography (see (Maier-Hein et al., 2017)).

| **Training set** | | **Results when tracking in the WM mask** | | | | |
|---|---|---|---|---|---|---|
| **Name** | **VS** | **VB** | **VS** | **F1** | **OL** | **$OR_{gt}$** |
| Prob (No IS) | 100.0 | 21 | 92.6 | 67.9 | 61.0 | 16.3 |
| IS1 | 95.2 (↓ 5%) | 21 | 88.9 (↓ 4%) | 67.8 (↓ 0%) | 60.8 | 16.3 |
| IS2 | 90.9 (↓ 10%) | 21 | 85.6 (↓ 8%) | 66.6 (↓ 2%) | 58.4 | 14.7 |
| IS3 | 83.3 (↓ 17%) | 21 | 81.1 (↓ 12%) | 65.7 (↓ 3%) | 56.9 | 13.9 |
| IS4 | 66.7 (↓ 33%) | 21 | 71.9 (↓ 22%) | 63.3 (↓ 7%) | 53.5 | 11.7 |

Experiment 5: Results with in vivo data. At the time of running experiment 5, we gained access to a larger (80 GB) GPU node from Alliance Canada clusters (NVidia H100 SXM5), which allowed us to use bigger batches (100 streamlines per batch). Both models were trained with a patience of 15 epochs without improvement, for a total of 65 epochs (i.e. 3 days; 45 hours of training and 28 hours of validation) for Learn2track, and 61 epochs (i.e. 6 days; 79 hours of training and 70 hours of validation) for TractoTransformers. To avoid loading volumes at each new batch, we used only 5 subjects per batch, and the same subjects for 25 batches before loading new volumes. To account for the larger size of the database, learning rates were generally smaller than in previous experiments. Visual inspection of results is showcased in **Figures 10 to 12.** Overall, our models could reproduce most bundles, even in subject sub-1222, which belongs to the site rejected from the training set, indicating good generalization capabilities of the models. However, even though the ratio of valid streamlines could not be evaluated here (see more on this in the discussion), **Figure 13** shows that it is probably higher than in the ISMRM data.

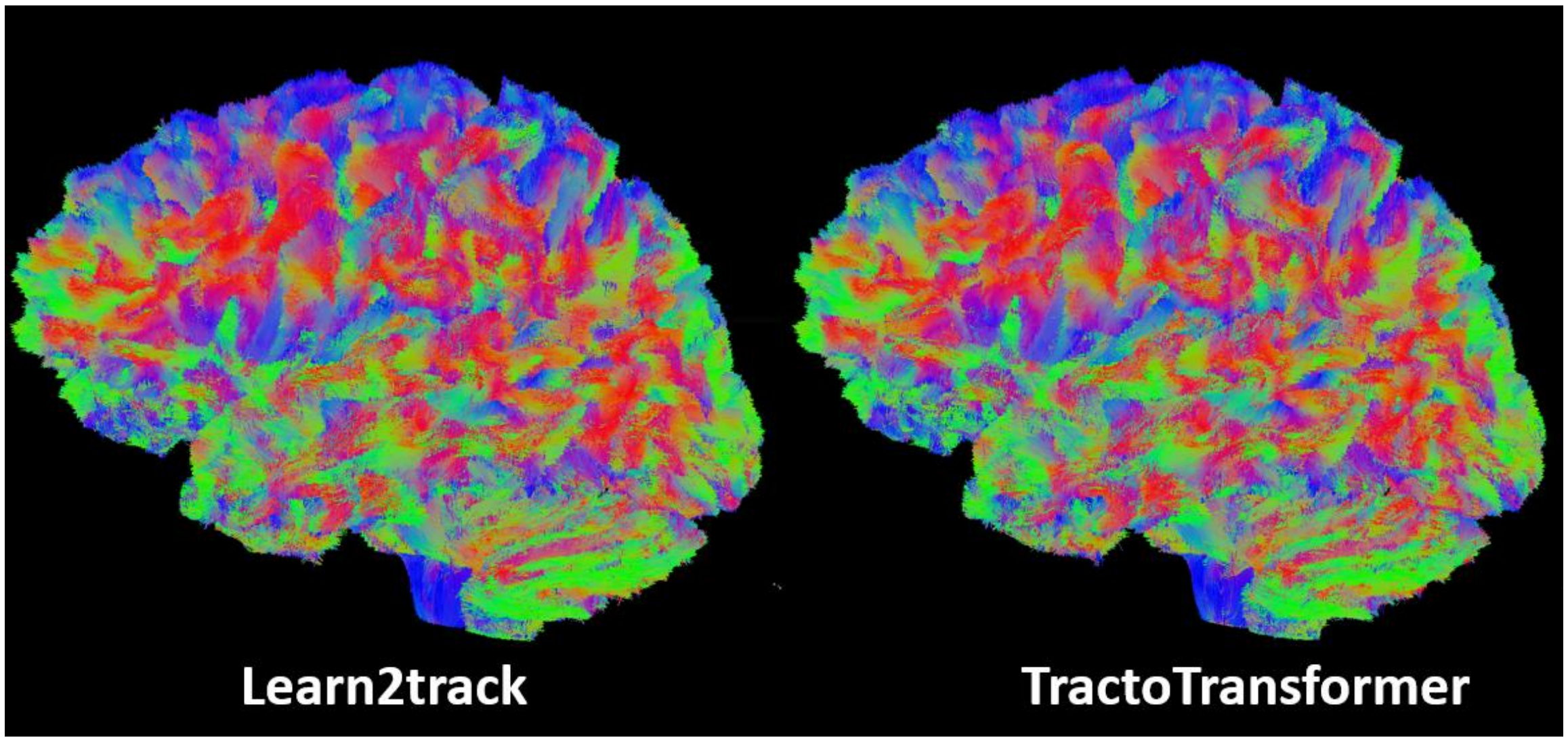

**Figure 10.** Experiment 5: Examples of wholebrain results on subject sub-1178. Visually, results seem to cover all gyri in the brain. Further analyses require bundle segmentation.

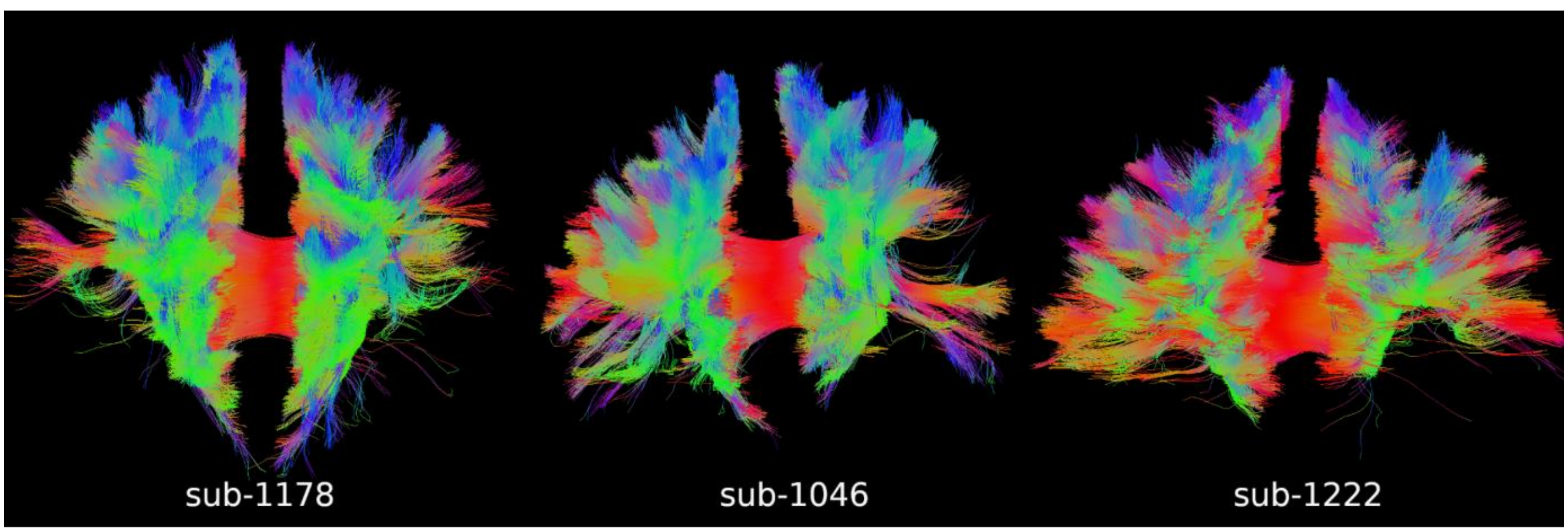

**Figure 11.** Experiment 5: results on all three testing subjects with the final Learn2track model: the CC_Fr_1 bundle segmented with BundleSeg (St-Onge et al., 2023) with a threshold of 10, anterior view.

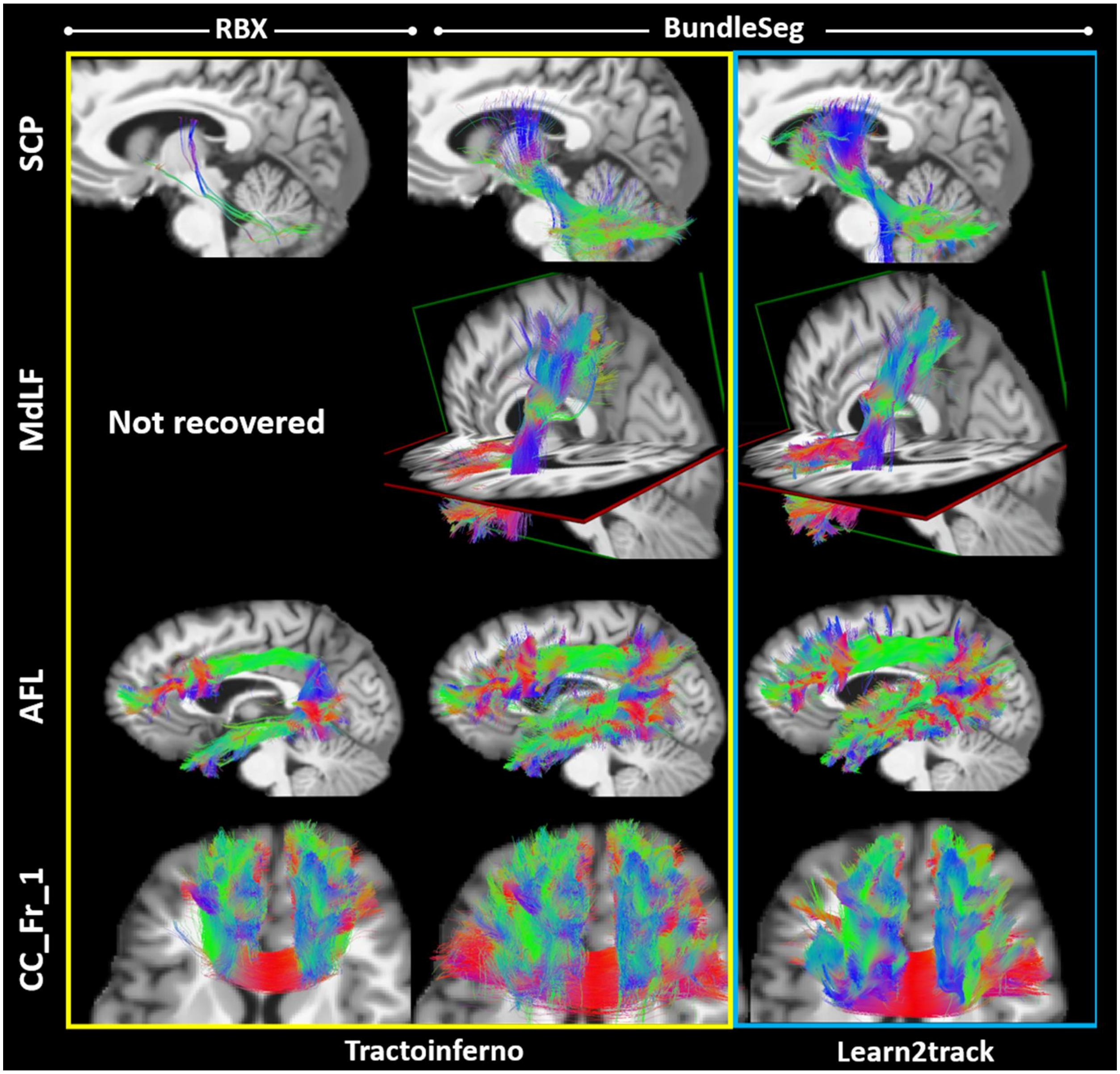

**Figure 12**. Experiment 5: Visual analysis of results at test time for subject sub-1046. SCP: left view. AFL: left view. CC_Fr_1: superior view. Yellow box: Tractoinferno's results, either with RBX as in the original publication (left) or with BundleSeg (right). Blue box: Learn2track's results when using BundleSeg. With BundleSeg, Learn2track's resulting bundles are similar to Tractoinferno's, even though the model was trained on training subjects' RBX bundles, which are generally stricter.

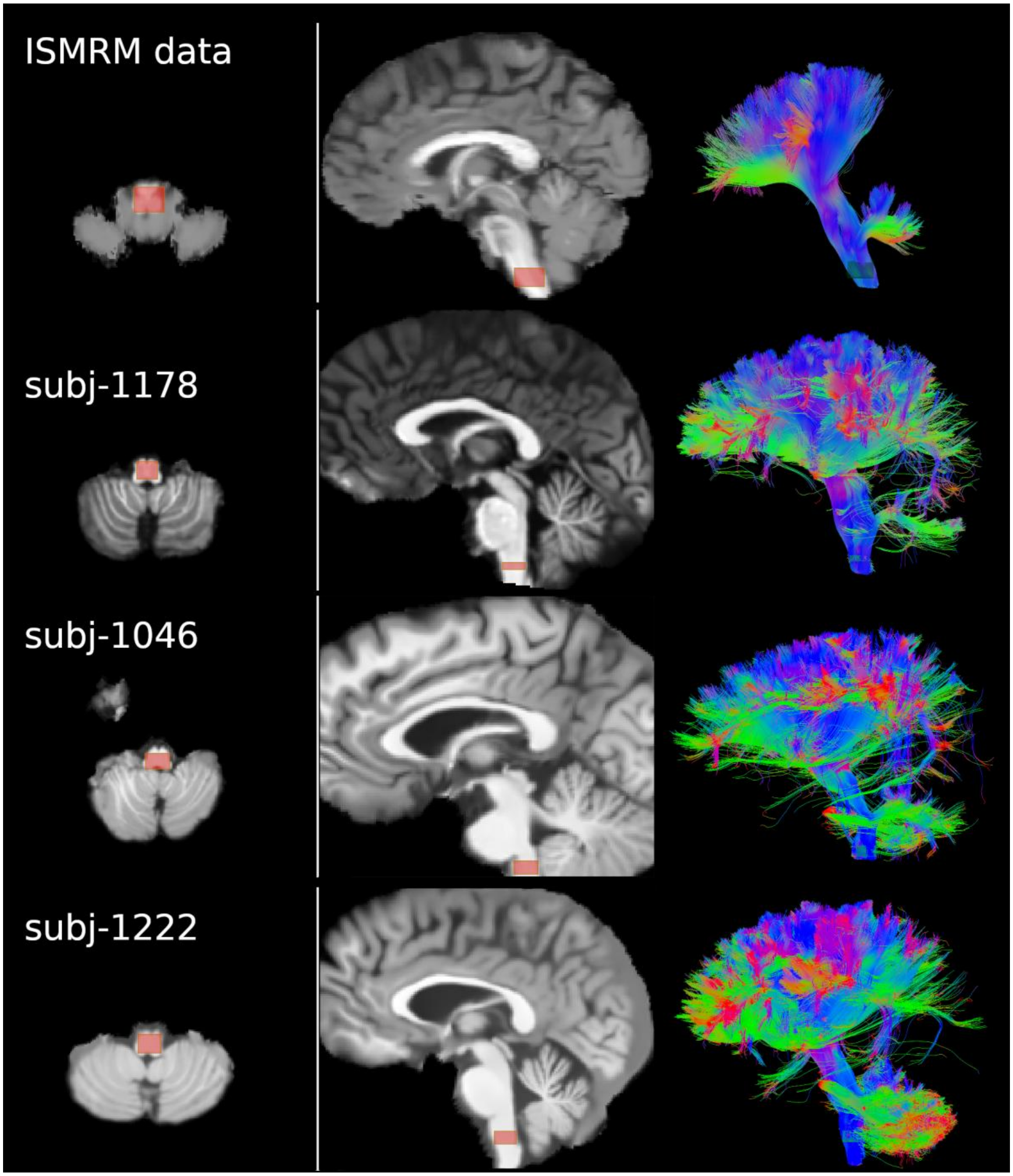

**Figure 13.** Comparison of *in vivo* results (Experiment 5) with results on the ISMRM data (Experiment 1). Streamlines passing through a small region of interest (ROI) manually positioned in the brainstem. In the ISMRM data, only valid streamlines are selected with this technique (here we can see the brainstem projection system [BST] and inferior cerebellar peduncle [ICP] bundles), whereas in Tractoinferno subjects, we can see spurious streamlines, and streamlines changing pathway to continue in the middle cerebellar peduncle (MCP) trajectory. Although we cannot measure the ratio of invalid streamlines with *in vivo* data, here we can see that results on the Tractoinferno database were of lesser quality than results on the ISMRM data.

Visualisation of the attention of Transformers. Visualisation of the Transformer attention in a selection of streamlines is shown in **Figure 14**. Here, we show attention weights for the first layer, for the average of the two heads. We can see that the attention spans on different points throughout time, and that it does not focus only on the current point. The map of the average weight is shown in **Figure 15**. This time, we show the average of the three layers. We can see that some voxels receive on average higher attention weights. Bertviz visualisation is available in **Supplemental HTML file 1**.

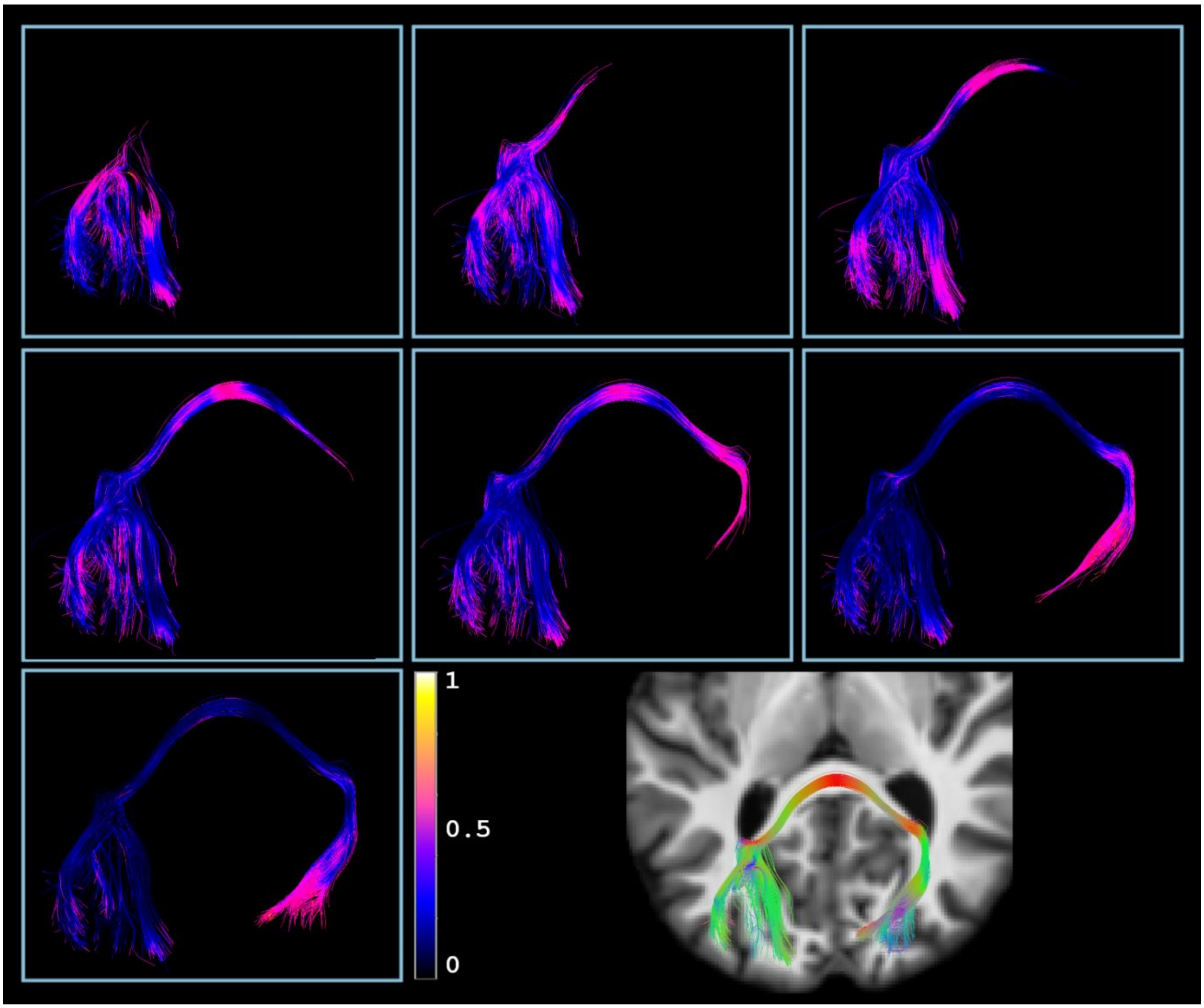

**Figure 14**. Visualisation of the attention of the Transformer during the tracking process of a section of the occipital corpus callosum (CC) in subject sub-1178. Colors here represent the average attention of the two heads of the first layer. Colors are rescaled for a better visualization (see description in **Supplemental Figure S2**). Colors on the last image (bottom, right) represent the local orientation of segments (red: left-right, green: anterior-posterior, blue: inferior-superior). The video of the full process is available as **Supplemental Video 1.**

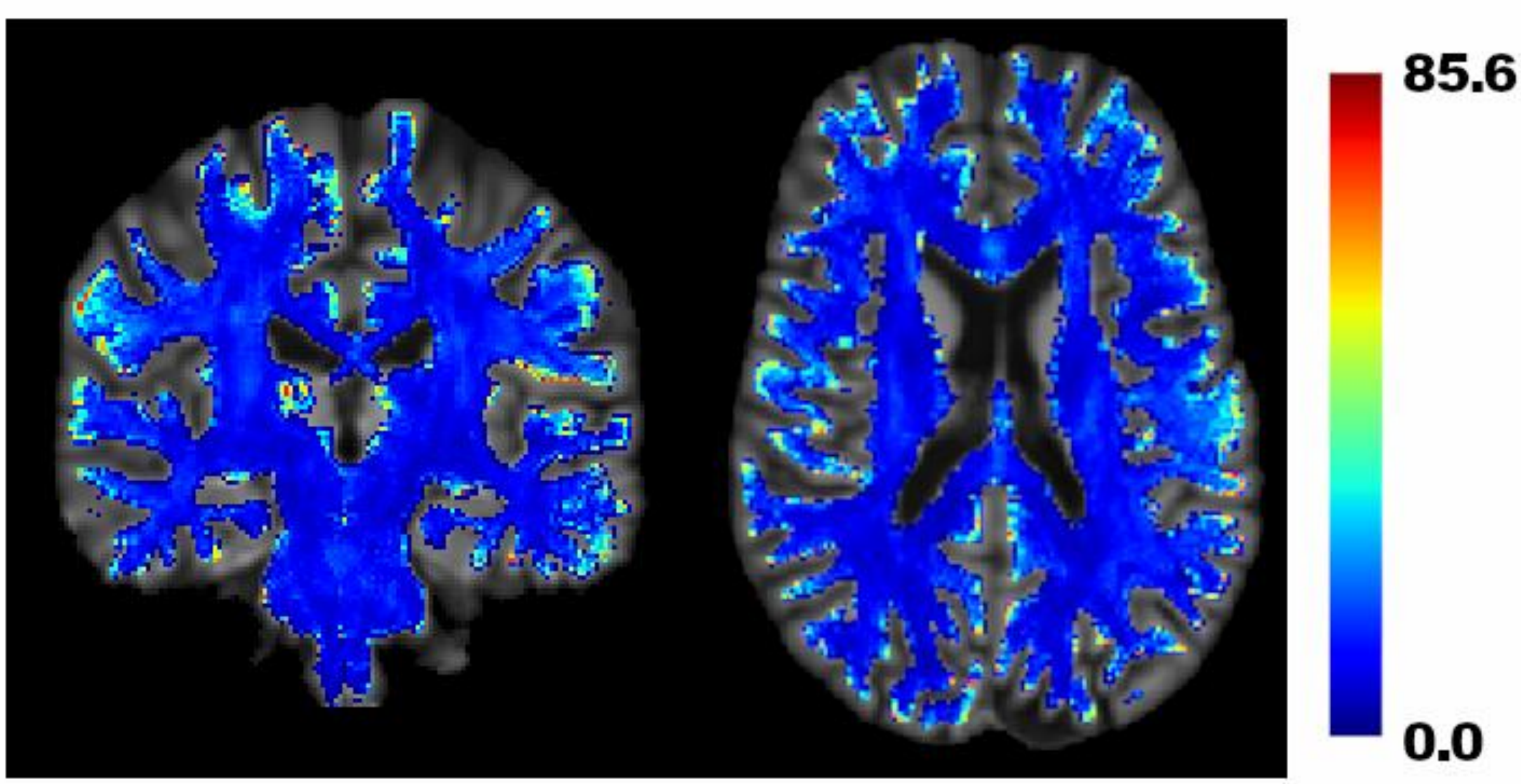

Figure 15. Map of the average weight of the Transformer (from experiment 5) at each voxel for subject sub-1178.

## 4. Discussion

We have successfully trained RNN and Transformer models for the task of tractography. These impressive results can be explained by careful supervision through a practical evaluation strategy, the GV phase, and by an extensive hyperparameter search. To the best of our knowledge, our results by far outperform previous state of the art methods on the ISMRM data.

To address the discrepancy between the local loss and the global tracking ability of our models, at validation time supervision was based on global metrics, even though the models were trained using a local loss. We used a generation-validation (GV) phase in which the first *n* points of validation streamlines were provided, and the models were used to complete the streamlines through tractography. The generated streamlines were then scored. The chosen metric needed to be computable on batches of streamlines (i.e., not requiring whole-brain tractography) and applicable to *in vivo* data (i.e., not relying on ground-truth comparisons). In this work, we used the false positive rate based on the connectivity matrix, for each generated batch. Many other metrics could have been considered, such as the distance between the endpoint of a generated streamline and that of its reference streamline in the validation set. However, we did not want to penalize models that generated streamlines different from their references, as long as the generated streamlines were valid. Using the connectivity matrix scores high any acceptable variation, but comes with the limitation that binary connectivity matrices with *in vivo* data are difficult to obtain and require heavy pre-filtering of the reference tractograms (Sotiropoulos & Zalesky, 2019; Yeh et al., 2021). In this work, we only used RBX-segmented bundles in the training set and when creating reference connectivity matrices, and these bundles are conservative (see **Figure 12** and discussion below).

Overall, Learn2track models performed better than TractoTransformer models and are faster at tractography time.

However, Transformers offer the advantage that the weights of the attention layers can be visualized, which could potentially support explainable AI. This process is sometimes critiqued (Jain & Wallace, 2019), but appears to offer the promise of providing insights on the input data (Wiegreffe & Pinter, 2019), which could potentially support explainable AI and improve our understanding of how to formulate anatomical priors in classical tractography. Yang et al. (2025) had already presented visualisation of the weights of their model, but they used matrices, which only enabled visualising one streamline at the time. The map of the average of weights such as presented in **Figure 15** could help explore the model's interpretability but is still difficult to analyse considering that voxels all contain a different number of streamlines, which limits interpretation. In **Figure 15,** voxels on the extremities appear more important to the Transformer, but this could be due only to the impact of the averaging step. The new coloring of tractograms presented in **Figure 14** could facilitate interpretation of the weights, and, although this still relies heavily on careful inspection and expert interpretation and is difficult to analyze on full-brain tractograms containing millions of streamlines, provides a way to view more data at once.

We introduce key takeaways:

Hyperparameter search is critical: With experiment 1, we showed that hyperparameter choices can generate major variations in the final quality of trained models. Here, we discuss a few interesting parameters.

- In our work, the addition of EOS had a particularly important influence on the results. This parameter is interesting because it removes the use of a tracking mask during tractography. Although some authors had mentioned that the stopping criterion could be learned (Dai et al., 2023; Poulin et al., 2017), to our knowledge it had only been implemented once (Waizman et al., 2025). Its importance should be highlighted.

- We observed similar conclusions to that of Poulin et al. (2022), namely that some loss functions are easier to train than others. Classification models were faster to train, but regression models led to better results. Gaussian and Fisher-von-mises models were less stable during training. Note that Fisher-von-mises models here did not integrate later work from Wegmayr and Buhmann (2020) such as entropy regularization.

- The step size (the distance between points of the streamline after resampling) also had a strong influence. A smaller step size implies longer sequences. It is a known limitation of RNNs that long-term dependencies are difficult to capture. For Transformers, longer sequences increase the size of the attention mechanism, which can dilute attention across many points and make learning more challenging. This parameter is important because during tractography (at generation time), the step size of tractography is also fixed to the same value. The initial step size was set to 1mm, similar to what is done typically in tractography, but we observed that increasing the step size to 2mm helped the models without decreasing tractogram quality.

- The choice of features in the input volume matters, but in a limited fashion. Adding the previous directions as input had limited effect. This probably shows that models were already able to have a good internal representation of the data without it. It could also be due to the use of wholebrain tractograms in the training set. For intuition, let's imagine a model training on bundles presenting a generally linear trajectory, and let's suppose that the model has learned that the next direction should be identical to the previous direction. Because streamlines are generally smooth, the loss could remain low: the angle between two segments of a streamline is rarely very large. However, during generation, the model would ignore the dMRI inputs and produce only straight lines. This drawback is not often possible in other fields of studies: in language models, the order of words in a sentence varies more than the order of segment orientations in a streamline. In music models, other specific patterns can be noticed, which has also required model adaptations (Huang et al., 2018). Here, using wholebrain data as well as limiting the inputs to MRI volumes could be important, highlighting again the importance of using type-3 models (see **Figure 1**). Concerning the input volumes, concatenating the T1 and FA to the fODF was helpful, but moderately. There have been some cases, in the literature, of models trained directly on the raw (denoised) dMRI. This has the advantage of allowing to skip many steps classically included in preprocessing pipelines, such as the computation of the local model, for instance using DTI, ODF or fODF. However, it has the disadvantage that each acquisition scheme comes with its own 4D volume format, including varying number of values per voxel and signification of each value. To allow using raw data with a multiple-site, multiple-acquisition scheme database, one needs to resample the raw dMRI to a common scheme on the sphere, which is not straightforward. Here on the ISMRM data, we did not observe that training on the raw data offered better results. Of note, other choices of inputs have been tested in the literature (see for instance Liu et al. (2025) or Zhao et al. (2023)) and have not been tested here. This should be included in future work.

- There have been many suggestions of models including CNN in the literature (Liu et al., 2025; Waizman et al., 2025; Yang et al., 2025). Here, adding CNNs as an embedding layer did not lead to an improvement of results as compared to NN embedding. In our models, at each point, the neighborhood is centered around the current position. The strength of CNNs is that they allow feature detection invariant to translation. But here, the features position in the field of view (the neighborhood) might be exactly what the model requires to estimate the next direction, and the model might not need a translation-invariant process.

The local loss is enough, but is not representative of the model's quality: Experiment 1 confirmed our hypotheses: training models with a local loss is effective, as evidenced by the obtained scores, but remains limited, as indicated by its weak correlation with other evaluation metrics. Many local losses were tested in our work, but they all used comparison to the direction of reference streamlines in the training set. Another option would be to use comparison with the direction of peaks in the dMRI data at each underlying point of the reference streamlines. Both choices are valid but come with their own limitations.

The main limitation of option 1 (comparing the output with the reference direction) is that training could be prone to overfitting on the shape of streamlines in the training set. In experiment 1, in most cases, the GV phase's connectivity FP rate stopped improving even when the training using the local loss was still decreasing. We also observed overfitting of the local loss on the training set in many cases, even though we investigated the effect of regularization techniques such as weight decay, dropout or noise addition. However, given that our models are used for generation, which is prone to the exposure bias as explained before, overfitting is not necessarily bad. Here, overfitting means that models learn to produce exactly the training set's directions when provided with exactly the training set's inputs. During generation, it is unclear whether the produced streamlines will look like streamlines from the training set or not, and even if they do, it might not lead to low scores as the produced streamlines will be valid.

The main limitation of option 2 (comparing the output with reference dMRI peaks) is that good tractograms (e.g., those with high F1 scores) are generally not associated with good alignment to local peaks (Schilling et al., 2019). Consistent with this, Experiment 3 revealed AE scores corresponding poorly to the F1, and that deterministic tracking leads to stronger AE scores than probabilistic tracking, but with lower F1. Using reference dMRI peaks would likely lead to deterministic-like results, no matter the choice of training set.

In short, training on the local loss appears as a possible solution, but limited. Other solutions have been suggested in the literature, such as sequence-level learning (Ranzato et al., 2016), where the model generates a full sequence, which is scored with a reward function. In the case of the ISMRM data, most scores are tractogram-based, not streamline-wise scores, but we can score individual streamlines as valid or invalid. In *in vivo* data, a similar score can be achieved by comparing with a gold standard connectivity matrix, such as achieved in this work during the GV phase. During optimization, the model is updated to increase probability of high-reward sequences. Of note, sequence-based learning is unstable, and the model must be initialized with a local loss training. Other possible solutions include scheduled sampling (Bengio et al., 2015; Mihaylova & Martins, 2019), where ground-truth inputs are gradually replaced with model-generated inputs during training, or minimum risk training (Shen et al., 2016), which includes evaluation metrics in the training process. To our knowledge, such sequence-level learning techniques have not yet been tested for the tractography task and could be the focus of future work.

Good supervision would require tracking at each epoch, and the GV phase is a good alternative: If the local loss is not a good indicator of success, then hyperparameter validation must be supervised with full tractogram generation, at each epoch. However, this is time-consuming. Our GV phase allowed quicker supervision. It also presents the advantage that models can be compared on-the-fly during training even if they use different local loss functions, which is useful when considering that models can take weeks to train, particularly on large databases such as Tractofinferno. In the case of the ISMRM data, experiment 1 showed good correlation between the FP rate of the connectivity matrix and the final tractometry score, even considering limitations in the preparation of the connectivity matrix.

Our models reproduce the structure of the training set rather than to explore novel anatomical pathways: Experiment 2 showed that our models were not able to explore the brain and to produce streamlines that they had not seen during training, or very poorly. This inability to track bundles not present in the training set explains the high valid streamline (VS) scores and limits the generation of anatomically implausible bundles. However, it also highlights the importance of carefully curated datasets when working with *in vivo* data. Experiment 3 showed that bundles presented smaller fanning in their extremities when trained on deterministic streamlines as with probabilistic streamlines, again reproducing characteristics of the training set. Overall, the training set should contain every bundle that users hope to reproduce, and with good coverage. The power of machine learning models seems to reside not in their ability to teach us about the brain, but in their ability to drastically reduce the number of false positives / invalid streamlines. However, the training set should still be curated as much as possible. Experiment 4 showed that the ratio of invalid streamlines increases when invalid streamlines are added to the training set.

One could also suggest that bundle-specific tractography (BST) models may be easier to train than wholebrain models. However, we believe that training on a single bundle, where most streamlines look very similar, is dangerous in terms of overfitting. Also, there is a possibility that models discover some important features when looking elsewhere in the brain. Here in Experiment 2, scores were lower for all bundles when removing a bundle from the training set, not just for that bundle. For instance, when removing the CC, scores in the CC were evidently low, but scores in other bundles also were lower.

However, BST should be used to prepare higher-quality training sets. Here, for *in vivo* data, the Tractoinferno database was used, which generated tractograms with Ensemble Tracking (Takemura et al., 2016), combining deterministic, probabilistic, particle filtering (Girard et al., 2014) and surface-enhanced tractography (St-Onge et al., 2018) to produce dense tractograms. Instead, new training sets could be produced with BST to create carefully curated bundles. This is a time-consuming task that requires meticulous work based on expert knowledge from anatomists and from histology, dissection, or tracer studies (Thomas et al., 2014). But once such training sets exist, machine learning models could learn to reproduce these results faster and with few false positives. Preparing such datasets is drastically different from preparing white matter bundle atlases: although both rely on anatomical knowledge, curated training sets aim for subject-specific accuracy rather than population-level representations. This direction should be a focus of future research.

We need more data associated with scoring systems: Not only do we need more training sets,

but we also need more datasets associated with a ground truth and with a scoring framework. The ISMRM data remains, to date, the only human-like simulated dataset with an established scoring system (Maier-Hein et al., 2017; Renauld et al., 2023). This data present limitations, but allows an unrivaled opportunity to compare tractograms with a high degree of precision in bundle segmentation. But we need more phantoms. More simulated datasets are available (ex., the 99 simulated brains dataset by Neher et al. (2021)), but are not associated with an official scoring system that includes a precise bundle segmentation technique.

We need specifically-designed phantoms that allow fair and controlled comparison across teams: The ISMRM data was offered in the context of a challenge aimed at evaluating complete pipelines, from raw data to tractograms. However, this setup makes it difficult to disentangle the contribution of individual parameters. Even the tractography task itself depends on many parameters, including the seeding mask, the number of seeds per voxel, and the tracking mask as well as its usage (e.g., are the endpoint of streamlines inside the mask, or are they just outside the mask?). Tracking parameters are typically not included as hyperparameters in ML studies, but they are also critical, as they impact the quality of tractogram reconstruction as much as the model's architecture does. In particular, with ISMRM data or with *in vivo* data, the tracking mask has a major impact on the scores. If a streamline has just one step too many, the streamline could end in the wrong gray matter parcel and become a false positive in connectivity analysis. Similarly, if all streamlines protrude just one millimeter at the end of a fanning bundle compared to a GT reference, the overreach score can increase dramatically. To alleviate this effect, we suggest that new phantoms should be created specifically for machine learning analysis, meaning that any parameter outside of the models' hyperparameters should be pre-defined. For instance, phantoms should be presented with an official tracking mask, and with indications as to if the endpoints should be inside the tracking mask or just outside of it. Similarly, they should be presented with a seeding mask and a gray matter parcellation.

In our work, we used the ground truth (GT) WM mask for experiments 1-4. Similarly, in all experiments, we used as seeding mask the GT interface mask, namely the voxels including an endpoint of the GT streamlines. Although this may have helped improve artificially our scores as compared to results of teams who participated in the ISMRM challenge and did not have access to that information, it also reduced one level of analysis in scores' interpretation. We suggest that future work could also use GT masks as tracking parameters.

We still lack precise evaluation methods for in vivo data: Evaluating results in *in vivo* data still is difficult. Evaluation usually requires segmenting bundles, and existing techniques (Recobundles (Garyfallidis et al., 2018), BundleSeg (St-Onge et al., 2023) or machine-learning based segmentation for instance) involve parameters that can influence results. Although bundle segmentation in general has experienced many innovations throughout the years, these technologies still suffer from limitations. Those limitations may have limited impact for general bundle usage (e.g. connectomics analysis (Sotiropoulos & Zalesky, 2019) or tract profiling (Yeatman et al., 2012)), but they are *critical* in the context of tractogram comparison. We want to emphasize that current segmentation methods do *not* allow creating scores that are precise enough for comparison of machine learning models. For instance, recent papers in machine learning for tractography used TractoEval (https://github.com/scil- vital/TractoInferno) from the TractoInferno authors. This is based on Recobundles (Garyfallidis et al., 2018), which includes randomization during its process, for instance while grouping streamlines. This randomization depends amongst others on the ordering of streamlines in the Tractogram. For instance, when running TractoEval on the true Tractoinferno tractograms, we did not obtain 100% Dice scores, but rather ~80% and even lower. Another limitation of TractoEval is that it only includes scores for bundles that have passed quality control (QC). This means that hard-to-track bundles are not tested with TractoEval, and that the final Dice score only reflects the ability of the models to track easy bundles. It also means that individual bundles do not have the same weight in all subjects. Overall, we suggest that future papers should stop relying on TractoEval to compare results, as it is not robust enough nor precise enough to allow comparison. Also, comparison across subjects is not possible as the score of each subject is based on a varying number of bundles.

Other approaches include 1) using segmentation techniques that are not influenced by stochastic processes, 2) visual inspection. Segmentation using inclusion / exclusion criteria such as passage in regions of interest (ROI) is less dependent on random parameters, but few official techniques exist for this. Amongst others, we can name Xtract (Warrington et al., 2020), but it does not separate the tractography and segmentation steps and can't be used here, or TractQuerier (Wassermann et al., 2013), but it does not provide examples of standardized protocols to extract bundles. Other solutions include ML-based segmentation methods (Berto et al., 2020; Chen et al., 2021; Dumais et al., 2023; Gupta et al., 2017; Labra et al., 2017; Ngattai Lam et al., 2018), for which it is difficult to determine the level of randomness, and which are not yet validated by a general usage in the community (Théberge et al., 2026). In this context, visual inspection still appears as the only reliable analysis of *in vivo* tractogram, for which the segmentation process can be less precise. To benefit fully from visual inspection, the bundle segmentation technique should provide bundles that include as many streamlines as possible, as long as they fit the anatomical definition of the bundle. This allows visual inspection of the fanning of streamlines and of the general coverage capabilities of the model. Rejected streamlines cannot be considered as invalid: they could be anatomically valid, but simply not included in the bundle *such as defined by the segmentation technique.* In TractoEval, Recobundles was used with the default settings, and these parameters are quite strict, leading to conservative results (see **Figure 12**). In this work, we used BundleSeg (St-Onge et al., 2023), which has only one parameter as compared to three in Recobundles. We set this parameter as high as possible, based on visual appreciation of the resulting bundle. Difference in resulting bundles was major, highlighting that scores based on these segmentation techniques have poor anatomical meaning.

In short, we still lack a QC method offering quality metrics with a precision level that could allow comparing models to one another for *in vivo* databases (Legarreta et al., 2026).

Our models performed well *in vivo*: Notwithstanding previously discussed limitations, our models performed well on *in vivo* data. Based on visual inspection, we could see that the models generated fully formed bundles. When using BundleSeg with loose criteria, generated bundles were similar to bundles generated from TractoInferno's ensemble tracking. This is interesting considering that our models were trained on TractoInferno's RBX-segmented bundles, which are more restricted. This could be the result of training on more than one subject.

Overall, however, visual inspection revealed noisier results on *in vivo* data than on the ISMRM data, showcasing once again the need for more phantoms or for better supervision techniques.

Limitations to our work:

Considering that the ISMRM data contains only one subject, the same subject was used for the training set, validation set and testing set, which could lead to potential information leakage. There have been discussions as to the validity of this approach. Wegmayr et al. (Wegmayr & Buhmann, 2020) suggested that the only correct training set in this case is to use other *in vivo* data and hope that the model can generalize to ISMRM data during testing. However, we argue that the dMRI volume of the ISMRM data is not similar enough to real human brains to allow this, particularly in the context of testing hyperparameters. The ISMRM experiments should therefore not be interpreted as demonstrating generalization. Instead, these experiments provide a controlled benchmark in which the effects of model architecture, training strategy, loss function, and hyperparameters can be investigated while quantitative tractogram scoring remains available. The results primarily demonstrate the ability of the models to learn and reproduce the anatomical organization represented in the phantom. Evidence for generalization to new anatomical instances must instead come from experiments using independent subjects or phantoms with scoring techniques, which, as discussed above, lack. In our work, we did use the ISMRM data for hyperparameter transfer to *in vivo* experiments. This provided a form of evidence that the training strategy developed on the ISMRM phantom can be applied to *in vivo* data, but should not be interpreted as evidence of quantitative generalization. Because no equivalent ground-truth scoring framework is available for the *in vivo* data, the performance of the models cannot be assessed quantitatively in the same way as on the ISMRM phantom. The *in vivo* experiments should therefore be regarded primarily as a proof of feasibility, demonstrating that the approach can be transferred from the controlled phantom setting to real human data, rather than as a rigorous evaluation of generalization to unseen anatomical instances.

In our work, hyperparameter selection was performed based on iterative selection of parameters. When quantitative evaluation of the *in vivo* data becomes possible, more systematic approaches could provide a more rigorous model-selection procedure such as formal grid search or cross-validation.

Our work did not allow understanding why Transformers did not outperform RNNs in our case, contrary to the general literature. It is possible that our sequences were not long enough to have created a sufficiently strong advantage for attention over recurrence. Also, Transformers can be more sensitive to optimization parameters. This could be further investigated in future work.

## Conclusion

In this work, we have presented successful models for tractography in the ISMRM data or in *in vivo* data. This work systematically evaluated RNN- and Transformer-based models for iterative tractography, with careful analysis of training strategies, hyperparameter choices, and the impact of the local loss formulation. Our results highlight that, while sequence-based models better capture streamline dependencies than purely local approaches, the mismatch between local training objectives and global tractography quality remains a central limitation, notably due to exposure bias. Overall, these findings clarify the strengths and limitations of current ML-based tractography approaches and provide guidance for future work toward models that better integrate global anatomical consistency during generation.

## References

Attyé, A., Renard, F., Baciu, M., Roger, E., Lamalle, L., Dehail, P., Cassoudesalle, H., & Calamante, F. (2021). TractLearn : A geodesic learning framework for quantitative analysis of brain bundles. *NeuroImage*, *233*(February). https://doi.org/10.1016/j.neuroimage.2021.117927

Aydogan, D. B., & Shi, Y. (2021). Parallel Transport Tractography. *IEEE transactions on medical imaging*, *40*(2), 635-647. https://doi.org/10.1109/TMI.2020.3034038

Bayrak, R. G., Schilling, K. G., Greer, J. M., Hansen, C. B., Greer, C. M., Blaber, J. A., Williams, O., Beason-Held, L. L., Resnick, S. M., Rogers, B. P., & Landman, B. A. (2019). TractEM: Fast Protocols for Whole Brain Deterministic Tractography-Based White Matter Atlas. *bioRxiv*, (960), 651935. https://doi.org/10.1101/651935

Bengio, S., Vinyals, O., Jaitly, N., & Shazeer, N. (2015). Scheduled Sampling for Sequence Prediction with Recurrent Neural Networks. *Advances in Neural Information Processing Systems*, *28*. https://proceedings.neurips.cc/paper_files/paper/2015/hash/e995f98d56967d946471af29d7bf99f1-Abstract.html

Benou, I., & Riklin Raviv, T. (2019). DeepTract : A probabilistic deep learning framework for white matter fiber tractography. *Lecture Notes in Computer Science (including subseries Lecture Notes in Artificial Intelligence and Lecture Notes in Bioinformatics)*, *11766 LNCS*, 626-635. https://doi.org/10.1007/978-3-030-32248-9_70

Berto, G., Bullock, D., Astolfi, P., Hayashi, S., Zigiotto, L., Annicchiarico, L., Corsini, F., De Benedictis, A., Sarubbo, S., Pestilli, F., Avesani, P., & Olivetti, E. (2020). Classifyber, a robust streamline-based linear classifier for white matter bundle segmentation. *bioRxiv*, (2011).

Billot, B., Greve, D. N., Puonti, O., Thielscher, A., Van Leemput, K., Fischl, B., Dalca, A. V., & Iglesias, J. E. (2023). SynthSeg : Segmentation of brain MRI scans of any contrast and resolution without retraining. *Medical Image Analysis*, *86*, 102789. https://doi.org/10.1016/j.media.2023.102789

Chamberland, M., Scherrer, B., Prabhu, S. P., Madsen, J., Fortin, D., Whittingstall, K., Descoteaux, M., & Warfield, S. K. (2017). Active delineation of Meyer's loop using oriented priors through MAGNEtic tractography (MAGNET). *Human Brain Mapping*, *38*(1), 509-527. https://doi.org/10.1002/hbm.23399

Chen, Y., Zhang, C., Song, Y., Makris, N., Rathi, Y., Cai, W., Zhang, F., & O'Donnell, L. J. (2021). Deep Fiber Clustering : Anatomically Informed Unsupervised Deep Learning for Fast and Effective White Matter Parcellation. *International Conference on Medical Image Computing and Computer-Assisted Intervention*. http://arxiv.org/abs/2107.04938

Daducci, A., Dal Palù, A., Lemkaddem, A., & Thiran, J. P. (2015). COMMIT: Convex optimization modeling for microstructure informed tractography. *IEEE Transactions on Medical Imaging*, *34*(1), 246-257. https://doi.org/10.1109/TMI.2014.2352414

Dai, H., Bauer, M., Fletcher, P. T., & Joshi, S. (2023). Modeling the Shape of the Brain Connectome via Deep Neural Networks. *arXiv*. http://arxiv.org/abs/2203.06122

Dumais, F., Legarreta, J. H., Lemaire, C., Poulin, P., Rheault, F., Petit, L., Barakovic, M., Magon, S., Descoteaux, M., & Jodoin, P.-M. (2023). FIESTA : Autoencoders for accurate fiber segmentation in tractography. *NeuroImage*, *279*, 120288. https://doi.org/10.1016/j.neuroimage.2023.120288

Elaldi, A., Gerig, G., & Dey, N. (2023). *$E(3) \times SO(3)$-Equivariant Networks for Spherical Deconvolution in Diffusion MRI* (arXiv:2304.06103). arXiv. http://arxiv.org/abs/2304.06103

Garyfallidis, E., Côté, M. A., Rheault, F., Sidhu, J., Hau, J., Petit, L., Fortin, D., Cunanne, S., & Descoteaux, M. (2018). Recognition of white matter bundles using local and global streamline-based registration and clustering. *NeuroImage*, *170*, 283-295. https://doi.org/10.1016/j.neuroimage.2017.07.015

Girard, G., Whittingstall, K., Deriche, R., & Descoteaux, M. (2014). Towards quantitative connectivity analysis : Reducing tractography biases. *NeuroImage*, *98*, 266-278.

Gupta, T., Patil, S. M., Tailor, M., Thapar, D., & Nigam, A. (2017). BrainSegNet : A Segmentation Network for Human Brain Fiber Tractography Data into Anatomically Meaningful Clusters. *arXiv*. http://arxiv.org/abs/1710.05158

Hosseini, S. M. H., Hassanpour, M., Masoudnia, S., Iraji, S., Raminfard, S., & Nazem-Zadeh, M. (2022). CTtrack : A CNN+Transformer-based framework for fiber orientation estimation & tractography. *Neuroscience Informatics, Artificial Intelligence in Brain Informatics*, *2*(4), 100099. https://doi.org/10.1016/j.neuri.2022.100099

Hosseini, S.-M. H., Hassanpour, M., Masoudnia, S., Iraji, S., Raminfard, S., & Nazem-Zadeh, M.-R. (2022). Transformer-Based Framework for Fiber Orientation Estimation & Tractography. *Neuroscience Informatics*. https://doi.org/10.2139/ssrn.4157505

Huang, C.-Z. A., Vaswani, A., Uszkoreit, J., Shazeer, N., Simon, I., Hawthorne, C., Dai, A. M., Hoffman, M. D., Dinculescu, M., & Eck, D. (2018). Music Transformer : Generating music with long-term structure. *arXiv*, 1-14.

Jain, S., & Wallace, B. C. (2019). Attention is not Explanation. In J. Burstein, C. Doran, & T. Solorio (Éds.), *Proceedings of the 2019 Conference of the North American Chapter of the Association for Computational Linguistics : Human Language Technologies, Volume 1 (Long and Short Papers)* (p. 3543-3556). Association for Computational Linguistics. https://doi.org/10.18653/v1/N19-1357

Jbabdi, S., & Johansen-Berg, H. (2011). Tractography : Where Do We Go from Here? *Brain Connectivity*, *1*, 169-183. https://doi.org/10.1089/brain.2011.0033

Jelescu, I. O., & Budde, M. D. (2017). Design and Validation of Diffusion MRI Models of White Matter. *Frontiers in Physics*, *5*. https://doi.org/10.3389/fphy.2017.00061

Jones, D. K., Knösche, T. R., & Turner, R. (2013). White matter integrity, fiber count, and other fallacies : The do's and don'ts of diffusion MRI. *NeuroImage*, *73*, 239-254. https://doi.org/10.1016/j.neuroimage.2012.06.081

Joshi, A., Sharma, A., Goel, A., Jha, R. R., Ahuja, C. K., Bhavsar, A., & Nigam, A. (2024). Tract-RLFormer : A Tract-Specific RL Policy Based Decoder-Only Transformer Network. In A. Antonacopoulos, S. Chaudhuri, R. Chellappa, C.-L. Liu, S. Bhattacharya, & U. Pal (Éds.), *Pattern Recognition* (p. 258-275). Springer Nature Switzerland. https://doi.org/10.1007/978-3-031-78201-5_17

Labra, N., Guevara, P., Duclap, D., Houenou, J., Poupon, C., Mangin, J.-F., & Figueroa, M. (2017). Fast Automatic Segmentation of White Matter Streamlines Based on a Multi-Subject Bundle Atlas. *Neuroinformatics*, *15*(1), 71-86. https://doi.org/10.1007/s12021-016-9316-7

Legarreta, J. H., Petit, L., Rheault, F., Theaud, G., Lemaire, C., Descoteaux, M., & Jodoin, P. M. (2020). Tractography filtering using autoencoders. *arXiv*.

Legarreta, J. H., Schiavi, S., Tang, W., Banks, G., Cieslak, M., Schilling, K., De Luca, A., Tournier, J.-D., Kruper, J., Rheault, F., Sotiropoulos, S. N., Pestilli, F., Veraart, J., Yang, J. Y.-M., Descoteaux, M., Heilbronner, S., & Rokem, A. (2026). What needs to be standardized for reliable, reproducible, and robust tractography? *GigaScience*, giag034. https://doi.org/10.1093/gigascience/giag034

Li, H., Liang, Z., Zhang, C., Liu, R., Li, J., Zhang, W., Liang, D., Shen, B., Zhang, X., Ge, Y., Zhang, J., & Ying, L. (2021). SuperDTI: Ultrafast DTI and fiber tractography with deep learning. *Magnetic Resonance in Medicine*, *86*(6), 3334-3347. https://doi.org/10.1002/mrm.28937

Li, Y., Zhang, W., Zhu, X., Wu, Y., Rathi, Y., O'Donnell, L. J., & Zhang, F. (2025). DDTracking : A Deep Generative Framework for Diffusion MRI Tractography with Streamline Local-Global Spatiotemporal Modeling. *arXiv*. https://doi.org/10.48550/arXiv.2508.04568

Liu, W., Calixto, C., Warfield, S. K., & Karimi, D. (2025). Streamline tractography of the fetal brain in utero with machine learning. *Imaging Neuroscience*, *Volume 3*. https://doi.org/10.1162/imag_a_00537

Lu, Q., Liu, W., Zhuo, Z., Li, Y., Duan, Y., Yu, P., Qu, L., Ye, C., & Liu, Y. (2022). A transfer learning approach to few-shot segmentation of novel white matter tracts. *Medical Image Analysis*, *79*. https://doi.org/10.1016/j.media.2022.102454

Maier-Hein, K. H., Neher, P. F., Houde, J. C., Côté, M. A., Garyfallidis, E., Zhong, J., Chamberland, M., Yeh, F. C., Lin, Y. C., Ji, Q., Reddick, W. E., Glass, J. O., Chen, D. Q., Feng, Y., Gao, C., Wu, Y., Ma, J., Renjie, H., Li, Q., … Descoteaux, M. (2017). The challenge of mapping the human connectome based on diffusion tractography. *Nature*

*Communications*, *8*(1). https://doi.org/10.1038/s41467-017-01285-x
Mihaylova, T., & Martins, A. F. (2019). Scheduled sampling for transformers. *Proceedings of the 57th Annual Meeting of the Association for Computational Linguistics: Student Research Workshop*, 351-356.
Neher, P., Côté, M.-A., Houde, J.-C., Descoteaux, M., & Maier-Hein, K. (2017). Fiber tractography using machine learning. *NeuroImage*, *158*, 417-429. https://doi.org/10.1101/104190
Neher, P., & Maier-Hein, K. (2021). *Providing realistic ground truth and AI-ready data for fiber tractography : The 99 simulated brains dataset*. Proceedings of Annual Meeting ISMRM. https://cds.ismrm.org/protected/21MProceedings/PDFfiles/3406.html
Ngattai Lam, P. D., Belhomme, G., Ferrall, J., Patterson, B., Styner, M. A., & Prieto, J. C. (2018). TRAFIC: fiber tract classification using deep learning. *ProcSPIEIntSocOptEng*, 37. https://doi.org/10.1117/12.2293931
Peretzke, R., Maier-Hein, K., Bohn, J., Kirchhoff, Y., Roy, S., Oberli-Palma, S., Becker, D., Lenga, P., & Neher, P. (2023). *atTRACTive : Semi-automatic white matter tract segmentation using active learning* (arXiv:2305.18905). arXiv. http://arxiv.org/abs/2305.18905
Pestilli, F., Yeatman, J., Rokem, A., Kay, K., Takemura, H., & Wandell, B. (2014). LiFE : Linear Fascicle Evaluation a new technology to study visual connectomes. *Journal of Vision*, *14*(10), 1122-1122. https://doi.org/10.1167/14.10.1122
Poulin, P., Côté, M. A., Houde, J. C., Petit, L., Neher, P. F., Maier-Hein, K. H., Larochelle, H., & Descoteaux, M. (2017). Learn to track : Deep learning for tractography. *Lecture Notes in Computer Science (including subseries Lecture Notes in Artificial Intelligence and Lecture Notes in Bioinformatics)*, *10433 LNCS*, 540-547. https://doi.org/10.1007/978-3-319-66182-7_62
Poulin, P., Theaud, G., Rheault, F., St-Onge, E., Bore, A., Renauld, E., de Beaumont, L., Guay, S., Jodoin, P.-M., & Descoteaux, M. (2022). TractoInferno—A large-scale, open-source, multi-site database for machine learning dMRI tractography. *Scientific Data*, *9*(1), Article 1. https://doi.org/10.1038/s41597-022-01833-1
Ranzato, M., Chopra, S., Auli, M., & Zaremba, W. (2016). Sequence Level Training with Recurrent Neural Networks. *arXiv*. http://arxiv.org/abs/1511.06732
Renauld, E., Boré, A., Poirier, C., Valcourt-Caron, A., Karan, P., Théberge, A., Théaud, G., Edde, M., Poulin, P., Girard, G., Houde, J.-C., Gagnon, A., St-Onge, E., Little, G., Legarreta, J. H., Thoumyre, S., Grenier, G., Yamani, Z. E., Pineda, M. O., … Descoteaux, M. (2026). Tractography analysis with the scilpy toolbox. *Aperture Neuro*, *6*. https://doi.org/10.52294/001c.154022
Renauld, E., Théberge, A., Petit, L., Houde, J.-C., & Descoteaux, M. (2023). Validate your white matter tractography algorithms with a reappraised ISMRM 2015 Tractography Challenge scoring system. *Scientific Reports*, *13*(1), 2347. https://doi.org/10.1038/s41598-023-28560-w
Rheault, F., St-Onge, E., Sidhu, J., Maier-Hein, K., Tzourio-Mazoyer, N., Petit, L., & Descoteaux, M. (2019). Bundle-specific tractography with incorporated anatomical and orientational priors. *NeuroImage*, *186*, 382-398. https://doi.org/10.1016/j.neuroimage.2018.11.018
Sarwar, T., Seguin, C., Ramamohanarao, K., & Zalesky, A. (2020). Towards deep learning for connectome mapping : A block decomposition framework. *NeuroImage*, *212*(November

2019), 116654. https://doi.org/10.1016/j.neuroimage.2020.116654
Schilling, K. G., Daducci, A., Maier-Hein, K., Poupon, C., Houde, J.-C., Nath, V., Anderson, A. W., Landman, B. A., & Descoteaux, M. (2019). Challenges in Diffusion MRI Tractography—Lessons Learned from International Benchmark Competitions. *Magnetic resonance imaging*, *57*, 194-209. https://doi.org/10.1016/j.mri.2018.11.014
Schilling, K. G., Petit, L., Rheault, F., Remedios, S., Pierpaoli, C., Anderson, A. W., Landman, B. A., & Descoteaux, M. (2020). Brain connections derived from diffusion MRI tractography can be highly anatomically accurate—If we know where white matter pathways start, where they end, and where they do not go. *Brain Structure and Function*. https://doi.org/10.1007/s00429-020-02129-z
Schilling, K. G., Tax, C. M. W., Rheault, F., Landman, B. A., Anderson, A. W., Descoteaux, M., & Petit, L. (2022). Prevalence of white matter pathways coming into a single white matter voxel orientation : The bottleneck issue in tractography. *Human Brain Mapping*, *43*(4), 1196-1213. https://doi.org/10.1002/hbm.25697
Schmidt, F. (2019). Generalization in Generation : A closer look at Exposure Bias. *arXiv preprint*. http://arxiv.org/abs/1910.00292
Shen, S., Cheng, Y., He, Z., He, W., Wu, H., Sun, M., & Liu, Y. (2016). Minimum risk training for neural machine translation. In K. Erk & N. A. Smith (Éds.), *Proceedings of the 54th Annual Meeting of the Association for Computational Linguistics (Volume 1 : Long Papers)* (p. 1683-1692). Association for Computational Linguistics. https://doi.org/10.18653/v1/P16-1159
Smith, R. E., Tournier, J. D., Calamante, F., & Connelly, A. (2012). Anatomically-constrained tractography : Improved diffusion MRI streamlines tractography through effective use of anatomical information. *NeuroImage*, *62*(3), 1924-1938. https://doi.org/10.1016/j.neuroimage.2012.06.005
Smith, R. E., Tournier, J. D., Calamante, F., & Connelly, A. (2015). SIFT2 : Enabling dense quantitative assessment of brain white matter connectivity using streamlines tractography. *NeuroImage*, *119*, 338-351. https://doi.org/10.1016/j.neuroimage.2015.06.092
Sotiropoulos, S. N., & Zalesky, A. (2019). Building connectomes using diffusion MRI : Why, how and but. *NMR in Biomedicine*, *32*(4), e3752. https://doi.org/10.1002/nbm.3752
St-Onge, E., Daducci, A., Girard, G., & Descoteaux, M. (2018). Surface-enhanced tractography (SET). *NeuroImage*, *169*(November 2017), 524-539. https://doi.org/10.1016/j.neuroimage.2017.12.036
St-Onge, E., Schilling, K. G., & Rheault, F. (2023). BundleSeg : A Versatile, Reliable and Reproducible Approach to White Matter Bundle Segmentation. In M. Karaman, R. Mito, E. Powell, F. Rheault, & S. Winzeck (Éds.), *Computational Diffusion MRI* (p. 47-57). Springer Nature Switzerland. https://doi.org/10.1007/978-3-031-47292-3_5
Takemura, H., Caiafa, C. F., Wandell, B. A., & Pestilli, F. (2016). Ensemble Tractography. *PLOS Computational Biology*, *12*(2), e1004692. https://doi.org/10.1371/journal.pcbi.1004692
Theaud, G., Houde, J., Bor, A., Morency, F., & Descoteaux, M. (2020). TractoFlow : A robust , efficient and reproducible diffusion MRI pipeline leveraging Nextflow & Singularity. *NeuroImage*, *218*(March). https://doi.org/10.1016/j.neuroimage.2020.116889
Théberge, A., Descoteaux, M., Desrosiers, C., & Jodoin, P. M. (2021). Track-to-learn : A general framework for tractography with deep reinforcement learning. *Medical Image Analysis*, 102093. https://doi.org/10.1101/2020.11.16.385229

Théberge, A., Yamani, Z. E., Barakovic, M., Magon, S., Yang, J. Y.-M., Descoteaux, M., Rheault, F., & Jodoin, P.-M. (2026). BundleParc : Consistent white matter bundle parcellation without tractography. *Medical Image Analysis*, 104087. https://doi.org/10.1016/j.media.2026.104087

Thomas, C., Ye, F. Q., Irfanoglu, M. O., Modi, P., Saleem, K. S., Leopold, D. A., & Pierpaoli, C. (2014). Anatomical accuracy of brain connections derived from diffusion MRI tractography is inherently limited. *Proceedings of the National Academy of Sciences of the United States of America*, *111*(46), 16574-16579. https://doi.org/10.1073/pnas.1405672111

Tournier, J.-D., Calamante, F., & Connelly, A. (s. d.). *Improved probabilistic streamlines tractography by 2nd order integration over fibre orientation distributions*.

Vaswani, A., Shazeer, N., Parmar, N., Uszkoreit, J., Jones, L., Gomez, A. N., Kaiser, Ł., & Polosukhin, I. (2017). Attention is all you need. *Advances in Neural Information Processing Systems*, *2017-Decem*(Nips), 5999-6009.

Vig, J. (2019). A Multiscale Visualization of Attention in the Transformer Model. In M. R. Costa-jussà & E. Alfonseca (Éds.), *Proceedings of the 57th Annual Meeting of the Association for Computational Linguistics : System Demonstrations* (p. 37-42). Association for Computational Linguistics. https://doi.org/10.18653/v1/P19-3007

Waizman, I., Gusakov, Y., Benou, I., & Raviv, T. R. (2025, septembre 19). TractoTransformer : Diffusion MRI Streamline Tractography using CNN and Transformer Networks. *arXiv*. http://arxiv.org/abs/2509.16429

Wakana, S., Caprihan, A., Panzenboeck, M. M., Fallon, J. H., Perry, M., Gollub, R. L., Hua, K., Zhang, J., Jiang, H., Dubey, P., Blitz, A., van Zijl, P., & Mori, S. (2007). Reproducibility of quantitative tractography methods applied to cerebral white matter. *NeuroImage*, *36*(4), 1123-1138.

Warrington, S., Bryant, K. L., Khrapitchev, A. A., Sallet, J., Charquero-Ballester, M., Douaud, G., Jbabdi, S., Mars, R. B., & Sotiropoulos, S. N. (2020). XTRACT - Standardised protocols for automated tractography in the human and macaque brain. *NeuroImage*, *217*, 116923. https://doi.org/10.1016/j.neuroimage.2020.116923

Wassermann, D., Makris, N., Rathi, Y., Shenton, M., Kikinis, R., Kubicki, M., & Westin, C.-F. (2013). On Describing Human White Matter Anatomy : The White Matter Query Language. In K. Mori, I. Sakuma, Y. Sato, C. Barillot, & N. Navab (Éds.), *Medical Image Computing and Computer-Assisted Intervention – MICCAI 2013* (p. 647-654). Springer. https://doi.org/10.1007/978-3-642-40811-3_81

Wasserthal, J., Neher, P., & Maier-Hein, K. H. (2018). TractSeg—Fast and accurate white matter tract segmentation. *NeuroImage*, *183*, 239-253. https://doi.org/10.1016/j.neuroimage.2018.07.070

Wegmayr, V., & Buhmann, J. M. (2020). Entrack : Probabilistic spherical regression with entropy regularization for fiber tractography. *International Journal of Computer Vision*, *129*(3), 656-680. https://doi.org/10.1007/s11263-020-01384-1

Wegmayr, V., Giuliari, G., & Buhmann, J. M. (2019). Entrack : A Data-Driven Maximum-Entropy Approach to Fiber Tractography. *Lecture Notes in Computer Science (including subseries Lecture Notes in Artificial Intelligence and Lecture Notes in Bioinformatics)*, *11824 LNCS*(MI), 232-244. https://doi.org/10.1007/978-3-030-33676-9_16

Wegmayr, V., Giuliari, G., Holdener, S., & Buhmann, J. (2018). Data-driven fiber tractography with neural networks. *Proceedings - International Symposium on Biomedical Imaging*,

*2018-April*(Isbi), 1030-1033. https://doi.org/10.1109/ISBI.2018.8363747
Xiong, R., Yang, Y., He, D., Zheng, K., Zheng, S., Xing, C., Zhang, H., Lan, Y., Wang, L., & Liu, T. (2020). On Layer Normalization in the Transformer Architecture. *Proceedings of the 37th International Conference on Machine Learning*, 10524-10533. https://proceedings.mlr.press/v119/xiong20b.html
Xue, T., Zhang, F., Zhang, C., Chen, Y., Song, Y., Makris, N., Rathi, Y., Cai, W., & O'donnell, L. J. (2022). Supwma : Consistent and efficient tractography parcellation of superficial white matter with deep learning. *IEEE 19th International Symposium on Biomedical Imaging (ISBI)*. www.slicer.org
Yang, Y., Yuan, Y., Ren, B., Wu, Y., Feng, Y., & Zhang, X. (2025). Deep learning-based diffusion MRI tractography : Integrating spatial and anatomical information. *NeuroImage*, *317*, 121314. https://doi.org/10.1016/j.neuroimage.2025.121314
Yeatman, J. D., Dougherty, R. F., Myall, N. J., Wandell, B. A., & Feldman, H. M. (2012). Tract Profiles of White Matter Properties : Automating Fiber-Tract Quantification. *PLOS ONE*, *7*(11), e49790. https://doi.org/10.1371/journal.pone.0049790
Yeh, C.-H., Jones, D. K., Liang, X., Descoteaux, M., & Connelly, A. (2021). Mapping Structural Connectivity Using Diffusion MRI : Challenges and Opportunities. *Journal of Magnetic Resonance Imaging*, *53*(6), 1666-1682. https://doi.org/10.1002/jmri.27188
Zeng, R., Lv, J., Wang, H., Zhou, L., Barnett, M., Calamante, F., & Wang, C. (2022). FOD-Net : A deep learning method for fiber orientation distribution angular super resolution. *Medical Image Analysis*, *79*, 102431. https://doi.org/10.1016/j.media.2022.102431
Zhang, F., Xue, T., Cai, W., Rathi, Y., Westin, C.-F., & O'donnell, L. J. (2022). TractoFormer : A Novel Fiber-level Whole Brain Tractography Analysis Framework Using Spectral Embedding and Vision Transformers. *International Conference on Medical Image Computing and Computer-Assisted Intervention*. www.humanconnectome.org
Zhao, S., He, J., Li, Y., Zeng, Q., Feng, Y., Amelina, E., Amelin, M., & Xu, Y. (2023). Retinogeniculate Visual Pathway Reconstruction Using Reinforcement Learning. *2023 42nd Chinese Control Conference (CCC)*, 7935-7940. https://doi.org/10.23919/CCC58697.2023.10239952

## Ethics

This study exclusively used publicly available, de-identified datasets. No new data involving human participants were collected by the authors. According to institutional and national guidelines, the present secondary analysis of publicly available data does not require additional ethical approval.

## Data / code availability

Trained models and learned parameters of final experiments are available on Zenodo (10.5281/zenodo.19709562) and can be used with dwi_ml, release 1.1.1 (https://dwi-ml.readthedocs.io/).

## Author contributions

PP created the initial version of the code for Learn2track. ER reformatted the code, created the dwi_ml library, and added the code for Transformers. ER ran all experiments and wrote the paper. AT, HL and MD provided help through manuscript revision.

## Competing interests

The authors declare no competing interests.

## Acknowledgements

We thank the Digital Research Alliance of Canada (alliancecan.ca). ER thanks the Fonds de recherche du Québec—Nature et technologies (FRQNT) and the Natural Sciences and Engineering Research Council of Canada (NSERC) programs for funding this research. MD thanks the NSERC Discovery grants (RGPIN-202004818) and the Université de Sherbrooke institutional research chair in neuroinformatics.